\documentclass[10pt]{article} 

\usepackage[preprint]{tmlr}

\usepackage{amsmath,amsfonts,bm}

\def\eqref#1{equation~\ref{#1}}

\def\1{\bm{1}}

\DeclareMathAlphabet{\mathsfit}{\encodingdefault}{\sfdefault}{m}{sl}
\SetMathAlphabet{\mathsfit}{bold}{\encodingdefault}{\sfdefault}{bx}{n}

\newcommand{\R}{\mathbb{R}}

\newcommand{\KK}{\text{KK}}
\newcommand{\CFH}{\text{CFH}}

\newcommand{\ARI}{\text{ARI}}
\newcommand{\CH}{\text{CH}}
\newcommand{\Sil}{\text{Silhouette}}

\newcommand{\CoHiRF}{\texorpdfstring{CoHiRF}{CoHiRF}}
\newcommand{\CoHiRFb}{ \texorpdfstring{CoHiRF$_\mathcal{B}$}{CoHiRF_B}}

\newcommand{\RCoHiRF}{%
  \texorpdfstring{\ensuremath{\mathcal{R}\text{-CoHiRF}}}{R-CoHiRF}%
}

\newcommand{\RCoHiRFb}{\texorpdfstring{$\mathcal{R}$-CoHiRF$_b$}{R-CoHiRF_b}}

\newcommand{\Code}{\text{Code}}

\newcommand{\dset}{\mathcal{D}_n}

\def\bX{{\bf X}}
\def\bx{{\bf x}}
\def\Xq{\bf X_{q}}
\renewcommand{\R}{\mathbb R}
\newcommand{\Labels}{\mathbf{L}}

\newcommand{\vP}{\textbf{{P}}}
\newcommand{\K}{\textbf{{K}}}
\newcommand{\BCM}{\text{{BCM}}}
\newcommand{\GetClusters}{\texttt{{GetClusters}}}
\newcommand{\ChooseMedoids}{\texttt{{ChooseMedoids}}}
\newcommand{\UpdateParents}{\texttt{{UpdateParents}}}
\newcommand{\GetFinalLabels}{\texttt{{GetFinalLabels}}}

\newcommand{\TimeComplexity}[1]{\mathrm{Time}(#1)}

\newtheorem{theorem}{Theorem}

\usepackage{hyperref}
\usepackage{cleveref}
\usepackage{url}

\title{\CoHiRF{}: Hierarchical consensus for interpretable clustering beyond scalability limits}

\author{
\name K. Meziani\email meziani@ceremade.dauphine.fr \\
\addr CEREMADE, University Paris Dauphine-PSL, France
\AND
\name B. Belucci  \email bruno.belucci-teixeira@dauphine.eu \\
\addr CEREMADE, University Paris Dauphine-PSL, France
\AND
\name K. Lounici  \email karim.lounici@polytechnique.edu\\
\addr CMAP, École Polytechnique, France  
\AND
\name V. R. Kostic \email vladimir.kostic@iit.it\\
\addr Istituto Italiano di Tecnologia, Genova, Italy, \\
      University of Novi Sad, Serbia
}

\def\month{MM}  
\def\year{YYYY} 
\def\openreview{\url{https://openreview.net/forum?id=XXXX}} 

\usepackage{booktabs} 
\usepackage{subcaption}
\usepackage{multirow}
\usepackage{longtable}
\usepackage{graphicx}
\usepackage{amsmath,amssymb}
\usepackage{algorithm}
\usepackage{algpseudocode}

\usepackage{xcolor}
\usepackage{hyperref}
\usepackage{cancel}
\usepackage{stmaryrd}

\begin{document}

\maketitle

\begin{abstract}

We introduce \CoHiRF{} (Consensus Hierarchical Random Features), a hierarchical consensus framework that enables existing clustering methods to operate beyond their usual computational and memory limits. \CoHiRF{} is a meta-algorithm that operates exclusively on the label assignments produced by a base clustering method, without modifying its objective function, optimization procedure, or geometric assumptions. It repeatedly applies the base method to multiple low-dimensional feature views or stochastic realizations, enforces agreement through consensus, and progressively reduces the problem size via representative-based contraction.

Across a diverse set of synthetic and real-world experiments involving centroid-based, kernel-based, density-based, and graph-based methods, we show that \CoHiRF{} can improve robustness to high-dimensional noise, enhance stability under stochastic variability, and enable scalability to regimes where the base method alone is infeasible. We also provide an empirical characterization of when hierarchical consensus is beneficial, highlighting the role of reproducible label relations and their compatibility with representative-based contraction.

Beyond flat partitions, \CoHiRF{} produces an explicit Cluster Fusion Hierarchy, offering a multi-resolution and interpretable view of the clustering structure. Together, these results position hierarchical consensus as a practical and flexible tool for large-scale clustering, extending the applicability of existing methods without altering their underlying behavior.
\end{abstract}

We introduce CoHiRF (Consensus Hierarchical Random Features), a hierarchical consensus framework that enables existing clustering methods to operate beyond their usual computational and memory limits. CoHiRF is a meta-algorithm that operates exclusively on the label assignments produced by a base clustering method, without modifying its objective function, optimization procedure, or geometric assumptions. It repeatedly applies the base method to multiple low-dimensional feature views or stochastic realizations, enforces agreement through consensus, and progressively reduces the problem size via representative-based contraction.

Across a diverse set of synthetic and real-world experiments involving centroid-based, kernel-based, density-based, and graph-based methods, we show that CoHiRF can improve robustness to high-dimensional noise, enhance stability under stochastic variability, and enable scalability to regimes where the base method alone is infeasible. We also provide an empirical characterization of when hierarchical consensus is beneficial, highlighting the role of reproducible label relations and their compatibility with representative-based contraction.

Beyond flat partitions, CoHiRF produces an explicit Cluster Fusion Hierarchy, offering a multi-resolution and interpretable view of the clustering structure. Together, these results position hierarchical consensus as a practical and flexible tool for large-scale clustering, extending the applicability of existing methods without altering their underlying behavior.

\section{Introduction}
\label{sec:intro}
  
Real-world datasets exhibit a wide range of clustering structures, including approximately spherical clusters in standard benchmark datasets, non-convex structures in image segmentation \citep{spectralclustering2000}, manifold-like structures in single-cell genomics \citep{kiselev2019}, or density-connected patterns in spatial data \citep{campello2013}. Accordingly, the clustering literature offers a variety of methods, each designed to capture specific types of structure. For practitioners, this diversity naturally raises the question:
\textbf{"which method best suits my data?"}.\\

\textit{Partitional methods} such as K-Means \citep{macqueen1967} are well suited to
compact, well-separated clusters and remain widely used due to their computational
efficiency. Extensions into kernel spaces \citep{scholkopf1998,dhillon2004} elegantly handle non-linear boundaries, though at increased computational cost.

\textit{Density-based methods}, pioneered by DBSCAN \citep{ester1996density}, revolutionized clustering by discovering arbitrarily shaped clusters through density connectivity, without requiring the number of clusters to be specified. Subsequent refinements such as OPTICS \citep{ankerst1999optics} and HDBSCAN \citep{campello2013} further improve robustness and enable hierarchical structure discovery. However, their effectiveness diminishes in high-dimensional settings, where distance concentration affects neighborhood definitions \citep{beyer1999nearest}.

\textit{Graph-based methods}, particularly spectral clustering \citep{ng2001,vonluxburg2007}, capture manifold and non-linear structures by leveraging the spectral properties of graph Laplacians constructed from pairwise similarities. Subsequent developments, such as self-tuning spectral clustering\citep{zelnik2004} and scalable  approximations \citep{chen2011,wang2020}, have improved their practical applicability. SC-SRGF \citep{liu2020}, which we use as a representative spectral method, employs sparse random graph filtering to avoid explicit eigen-decomposition while preserving spectral information.

\textit{Hierarchical methods} \citep{murtagh2012} build dendrograms through
agglomerative or divisive strategies, providing multi-resolution descriptions of the data and offering valuable interpretability.


Each clustering method has its own strengths, shaped by decades of algorithmic innovation and is best suited to particular data structures. Yet, in practice, scalability with respect to both the sample size $n$ and the dimensionality $p$ remains a major challenge.

Hierarchical methods typically incur $O(n^2)$ or $O(n^3)$ complexity, which limits their applicability to moderate-scale datasets. Spectral clustering methods, including SC-SRGF, still face memory constraints for large $n$ due to the need to construct a similarity graph.
Density-based clustering methods, which rely on local density and neighborhood definitions, often struggle in high-dimensional spaces due to distance concentration and can become computationally expensive for large datasets. In low-dimensional settings, DBSCAN can be accelerated using spatial indexing structures such as kd-trees or ball trees, reducing the average complexity to
$O(n \log n)$; however, these structures become ineffective in high-dimensional spaces, leading to near-quadratic runtime.

Previous work has addressed scalability through several strategies. \textit{Sampling-based approaches} \citep{bradley1998,farnstrom2000} reduce the dataset size, but may lose important structural information, especially for rare or small clusters.
\textit{Dimensionality reduction} techniques \citep{boutsidis2010} project the data before clustering, which can accelerate algorithms such as K-Means, however the choice of target dimension and projection strategy remains challenging and the benefits for spectral or density-based methods are less systematic.
\textit{Approximation algorithms} \citep{kumar2004,bachem2016}, such as mini-batch
K-Means \citep{sculley2010} and accelerated variants \citep{elkan2003}, trade
solution quality for speed. These approaches primarily benefit centroid-based methods and remain largely method-specific,  they do not address the broader challenge of making other clustering families scalable.

As a result, the question faced by practitioners often shifts from \textbf{"which method best suits my data?"} to \textbf{"which method can actually be executed?"}. When both $n$ and $p$ are large, practical constraints substantially narrow the set of feasible clustering algorithms and K-Means frequently becomes the default choice. This prevalence does not stem from its ability to capture all types of structure, but from its favorable computational properties and consistent scalability.

This situation entails a tangible methodological limitation: datasets exhibiting non-spherical, density-based, or manifold structures are often analyzed using K-Means, even when other clustering paradigms would be more appropriate from a modeling perspective.

\subsection{Related work and motivation}
\label{sec:related}

Meta-clustering approaches have long been proposed to improve robustness and reliability by aggregating multiple partitions obtained from a base clustering method. \textit{Ensemble clustering} \citep{strehl2002,fred2005combining} and \textit{consensus clustering} \citep{monti2003,vega2007} methods aim to identify stable structures that persist across different runs, initializations, or algorithmic variants. While effective in small to medium-scale settings, the scalability of these approaches remains fundamentally limited by the computational and memory requirements of the underlying clustering methods.

\textit{Random projection–based ensemble strategies} provide an important step toward high-dimensional scalability. In particular, Random Projection Ensemble Clustering (RPEC) \citep{fern2003random} applies a base clustering method to multiple random projections and aggregates the results through a co-association matrix. RPEC demonstrates that random projections can mitigate high-dimensional noise while preserving clustering quality. However, the explicit construction of an $n \times n$ co-association matrix makes this approach impractical for large sample sizes.

Our motivation is to restore algorithmic choice in large-scale clustering. In practice, K-Means often dominates not because it best matches the data structure, but because it is one of the few clustering methods that reliably scales to large $n$ and high $p$. As a result, practitioners are frequently constrained to use centroid-based methods even when other clustering paradigms would be more appropriate from a modeling perspective.

We introduce \CoHiRF{} (\textbf{Co}nsensus \textbf{Hi}erarchical \textbf{R}andom \textbf{F}eatures), a hierarchical consensus meta-algorithm designed to enable existing clustering methods to operate beyond their usual computational limits. Rather than constructing a full co-association matrix, \CoHiRF{} relies on code-based consensus, hierarchical medoid contraction and batched processing. This design enables scalability with respect to both sample size and feature dimension, while producing an explicit hierarchical representation of cluster relationships.

\subsection{Contributions: \CoHiRF{} }
\label{sec:contribution}

\CoHiRF{} is \textbf{not} a new clustering algorithm. It is a meta-algorithm that operates on top of an existing Base Clustering Method (\BCM), without modifying its objective function, optimization procedure, or underlying assumptions, while extending its applicability beyond computational and memory limitations. \CoHiRF{} operates exclusively on label assignments through repeated consensus and hierarchical contraction. In doing so, it enables base clustering methods to operate at scales that would otherwise be inaccessible, by progressively reducing the problem size while retaining label relations that are consistently reproduced across runs.

\begin{figure}[ht]
\centering
\includegraphics[width=0.8\linewidth]{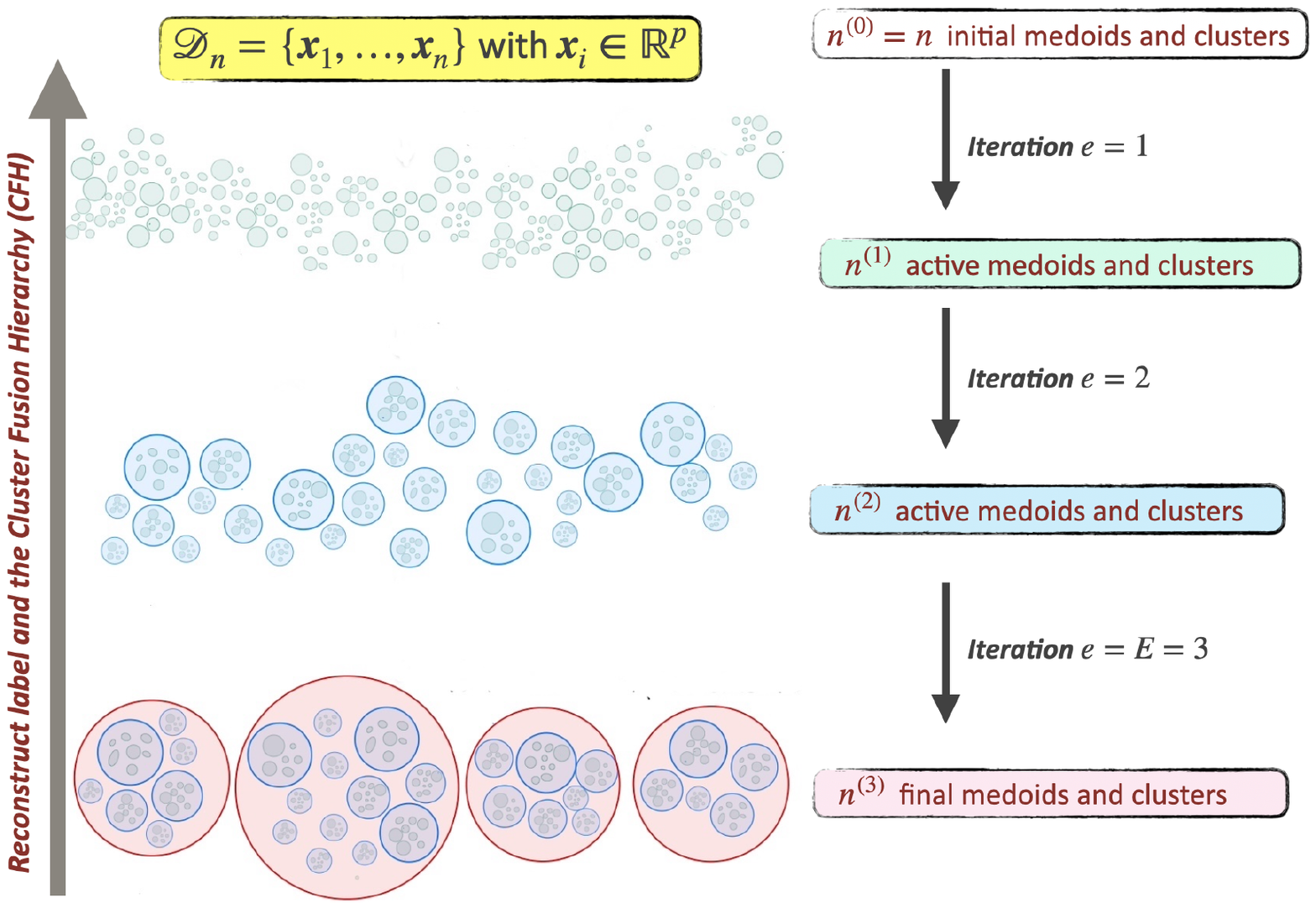}
\caption{\CoHiRF{} overview. The \BCM{}  is applied to multiple random feature views of the current active medoids. Stable label relations are identified through consensus, representative medoids are selected and the process is repeated until convergence, producing a Cluster Fusion Hierarchy (CFH).}
\label{fig:overview}
\end{figure}

The core principle of  \CoHiRF{} is to extract robust clustering structure through \emph{feature-based views}, without introducing any new notion of distance or similarity. The base clustering method is repeatedly applied to low-dimensional representations and  \CoHiRF{} identifies label relations that are remain consistent across views and progressively reducing the problem size through hierarchical contraction. Figure~\ref{fig:overview} provides an overview of the framework.


This process does not introduce new structure nor a new notion of proximity. Instead, it identifies label relations that are already produced by the base clustering method and remain reproducible across repeated applications of the method. The framework relies on three main mechanisms (Section~\ref{sec:method}):

\begin{itemize}
\item \textbf{Random feature views.}
At each iteration, $R$ subsets of $q$ features are sampled from the original $p$ dimensions, yielding low-dimensional representations on which the base method remains tractable. Applying the base method across multiple views probes the reproducibility of its label assignments. In some cases, these views arise from stochastic realizations of the base method rather than feature subsampling.
\item \textbf{Consensus.}
The base method is applied independently to each feature view. A strict consensus retains only label relations that are consistent across all views. A relaxed variant discards highly inconsistent views to improve robustness in noisy regimes.
\item \textbf{Hierarchical medoid contraction.}
Consensus clusters are summarized by representative medoids, reducing the effective problem size. Repeating this process yields a bottom-up hierarchy that progressively aggregates stable label relations.
\end{itemize}

The output of \CoHiRF{} is an explicit \textbf{Cluster Fusion Hierarchy (\CFH)}, which records all successive cluster fusions induced by consensus and hierarchical contraction. Unlike classical agglomerative dendrograms, the \CFH{} is not constructed from pairwise distance comparisons and is not restricted to binary merges. At each iteration, multiple groups may fuse simultaneously when they are consistently grouped together across repeated applications of the base clustering method. The resulting hierarchy therefore reflects consensus-driven cluster fusions obtained under controlled reductions of the problem size. As a result, the \CFH{} provides a multi-resolution representation of cluster relationships, without requiring the number of clusters to be specified in advance (Section~\ref{subsec:toy_iris}).

To scale to large sample sizes, \CoHiRF{} supports batch processing (Section~\ref{sec:batched}). The data is partitioned into batches on which the base method is applied independently. Representative medoids extracted from each batch are then aggregated hierarchically using the same consensus and contraction mechanisms. This strategy enables scalability in $n$ while remaining faithful to the behavior of the base method whenever stable label assignments are present.

The main contributions of this work are:

\begin{itemize}
\item[\textbf{(i)}] \textit{A general hierarchical consensus framework}
applicable to centroid-based, kernel-based, density-based and graph-based clustering methods.

\item[\textbf{(ii)}] \textit{A meta-algorithm that leaves the base clustering method unchanged},
operating exclusively on its label assignments without modifying its objective function,
optimization procedure, or geometric assumptions.

\item[\textbf{(iii)}] \textit{Scalability in both sample size and feature dimension}
through random feature views, hierarchical contraction and batch processing.

\item[\textbf{(iv)}] \textit{Strict and relaxed consensus mechanisms} exposing a robustness-flexibility tradeoff.

\item[\textbf{(v)}] \textit{An explicit, interpretable hierarchical output}
in the form of a Cluster Fusion Hierarchy (\CFH), which represents clustering structure at multiple resolutions instead of producing a single flat partition.

\item[\textbf{(vi)}] \textit{A theoretical characterization of computational and memory complexity}
for both standard and batched formulations.

\end{itemize}


The remainder of this paper is organized as follows.
Section~\ref{sec:method} introduces the \CoHiRF{} framework and its main algorithmic components.
Section~\ref{sec:overall_complexity} provides a theoretical analysis of the computational and memory complexity, including batched formulations.
Sections~\ref{sec:experiments_synthetic} and~\ref{subsec:Real_word_exp} present experimental results on synthetic and real-world datasets.
Section~\ref{sec:Empirical_insights} summarizes empirical observations on the behavior of hierarchical consensus across different base methods and data regimes.
Section~\ref{sec:conclusion} concludes the paper.

\section{The \CoHiRF{} framework}
\label{sec:method}

\subsection{Overview and notation}
\label{sec:algo_overview}

Consider a dataset $\dset = \{\bx_1, \ldots, \bx_n\}$ with $\bx_i \in \R^p$ and let $\bX \in \R^{n \times p}$ denote the data matrix. \CoHiRF{} builds a hierarchical clustering structure by iteratively grouping samples through consensus labeling and medoid selection.  The framework itself is agnostic to the notion of similarity or distance and relies exclusively on the labels produced by the base clustering method.

At iteration $e$, the algorithm operates on a set of \emph{active medoids}.
We denote by $\K^{(e-1)} \subseteq \llbracket n \rrbracket$ the index set of active medoids at iteration $e$, with cardinality $n^{(e-1)} = |\K^{(e-1)}|$ and by
$$
\bX^{(e)} = \bX[\K^{(e-1)}, \,:]
$$
the corresponding data matrix. Initially, all samples are active medoids, i.e., $\K^{(0)} = \llbracket n \rrbracket$. At each iteration, \CoHiRF{} applies three main functions:
\begin{enumerate}
\item \textbf{\GetClusters}: given the current active medoids $X^{(e)}$, this function applies the base clustering method across multiple random feature views and returns a consensus labeling
$
\Labels^{(e)} \in \llbracket n^{(e)} \rrbracket^{n^{(e-1)}}
$
of the active medoids (Figure~\ref{fig:Step1}).

\item \textbf{\ChooseMedoids}: for each consensus cluster, a representative medoid is selected. The indices of these medoids define the new active set $\K^{(e)}$, with cardinality $n^{(e)} = |\K^{(e)}|$, which forms the input for the next iteration (Figure~\ref{fig:Step2}).

\item \textbf{\UpdateParents}: the parent vector $\vP \in \llbracket n \rrbracket^{n}$ is updated for medoids that become inactive at the current iteration, recording the hierarchical relationships induced by cluster merges.
\end{enumerate}

\begin{algorithm}
\caption{\CoHiRF{}}
\label{alg:CoHiRF}
\begin{algorithmic}[1]
\Require $\bX \in \R^{n \times p},\, \BCM,\, R,\, q,\, \textrm{max\_iter}$
\Ensure Hierarchical structure $\{\bX^{(e)}\}_{e=0}^{E}$, Cluster Fusion Hierarchy (\CFH)
\State $\vP ,\,\K^{(0)}\gets (1:n),\,\,(1:n)$
\State $e,\, n^{(e-1)},\, n^{(e)}\gets 0,\, 0,\, n$
\While{$n^{(e-1)} \neq n^{(e)}$ \textbf{and} $e < \textrm{max\_iter}$}
    \State $\bX^{(e)} \gets \bX[\K^{(e-1)}, \,:\,]$
    \State $\Labels^{(e)} \gets \GetClusters\left(\bX^{(e)},\, \BCM,\, R,\, q\right)$
    \State $\K^{(e)} \gets \ChooseMedoids\left(\bX^{(e)},\, \Labels^{(e)}\right)$
    \State $n^{(e)} \gets |\K^{(e)}|$
    \State $\vP \gets \UpdateParents\left(\vP,\, \Labels^{(e)},\, \K^{(e-1)},\, \K^{(e)}\right)$
    \State $\K^{(e-1)},\, n^{(e-1)},\, e\gets \K^{(e)},\, n^{(e)}, e+1$
\EndWhile
\State $\Labels \gets \GetFinalLabels\left(\vP,\, \K^{(E)}\right)$
\State\Return $\{\bX^{(e)}\}_{e=0}^{E}$, \CFH{} encoded by $\vP$
\end{algorithmic}
\end{algorithm}

These three steps are applied iteratively, progressively reducing the number of active medoids as clusters merge at each iteration.   The algorithm terminates when no further reduction occurs, i.e., when $n^{(e)} = n^{(e-1)}$, or when a maximum number of iterations is reached.\\
Once convergence is achieved, the function \GetFinalLabels{} assigns a final cluster label to each original sample by propagating labels through the parent vector $\vP$. The parent vector naturally encodes the full hierarchy of successive cluster fusions. We refer to this hierarchical structure as a \textbf{Cluster Fusion Hierarchy (\CFH)}.

\subsection{Algorithm components}
\label{sec:algo_components}

\subsubsection{Consensus-based clustering (\GetClusters)}
\label{sec:GetClusters}

At iteration $e$, the function \GetClusters{} aggregates multiple clusterings of the current active medoids into a single consensus partition.
Let $\bX^{(e)} \in \R^{n^{(e-1)} \times p}$ denote the matrix of active medoids. The procedure performs $R$ independent clusterings using random feature views of $\bX^{(e)}$.

\begin{figure}[ht]
\centering
\includegraphics[width=0.8\linewidth]{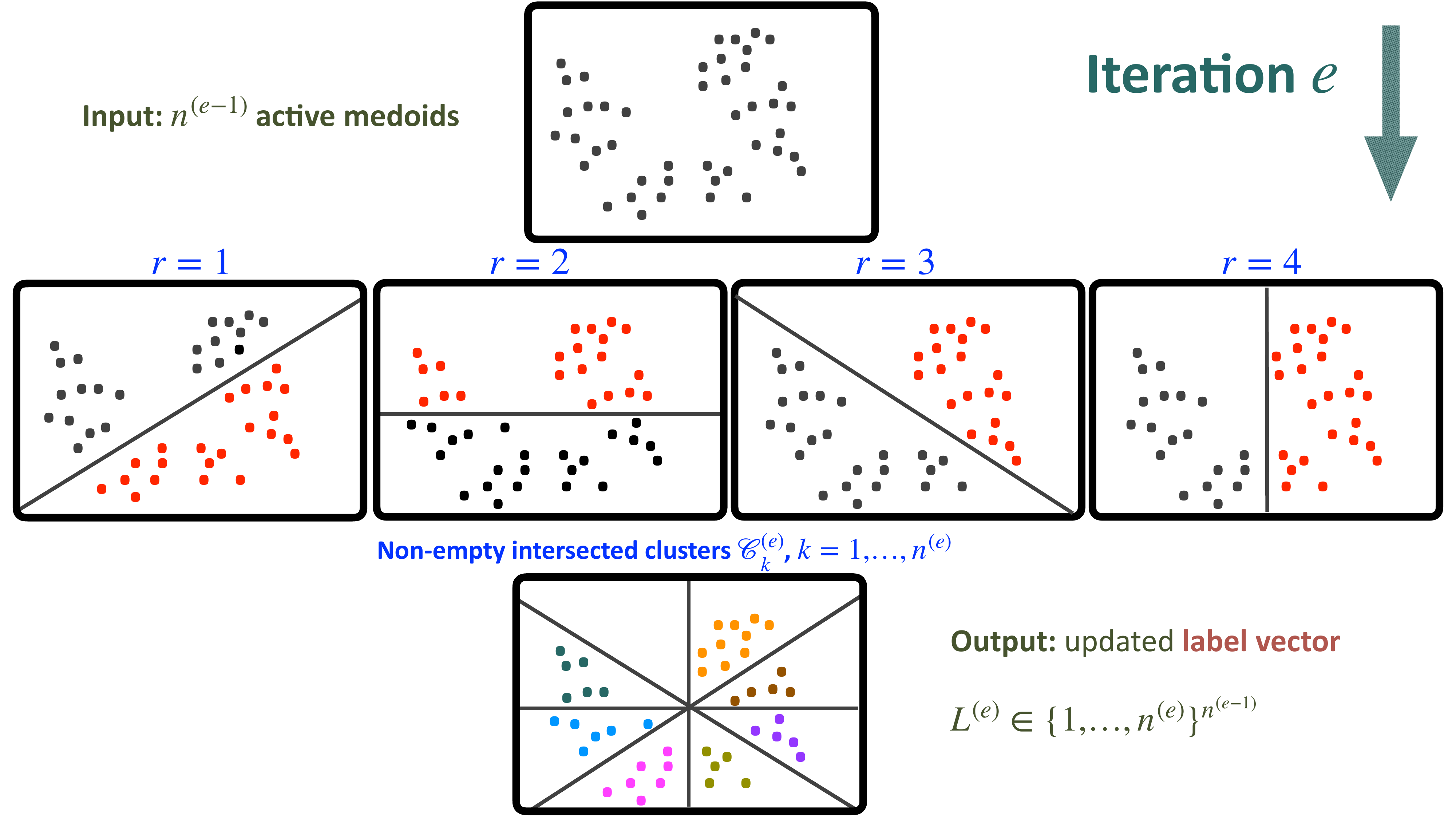}
\caption{\footnotesize Illustration of \GetClusters{} at iteration $e$: 
(i) random feature sampling, 
(ii) application of the \BCM{}  on each view, 
(iii) consensus formation across views.}
\label{fig:Step1}
\end{figure}

For each repetition $r \in \llbracket R \rrbracket$, a subset of $q$ features is sampled uniformly at random, yielding a projected data matrix
$\Xq^{(r)} \in \R^{n^{(e-1)} \times q}$. A base clustering method (\BCM{}, e.g., K-Means) is then applied to each projection, producing $R$ distinct partitions of the active medoids.
Figure~\ref{fig:Step1} illustrates this process.

Given $R$ partitions of the same set of $n^{(e-1)}$ samples, the goal is to derive a single partition that captures structures consistently observed across views.
While ensemble principles are well established in supervised learning, consensus construction in unsupervised settings is more challenging.
Direct approaches based on pairwise co-occurrence matrices or graph intersections typically require $O(R \, n^2)$ time and memory, which is prohibitive in large-sample regimes.

We address this issue through a \textbf{\emph{code-based strict consensus}} formulation.
For each active medoid $\bx_i$, we define a code that records its cluster assignments across all $R$ views:
\begin{equation*}
\Code(\bx_i) = (\ell_i^{(1)}, \ell_i^{(2)}, \ldots, \ell_i^{(R)}) \in \mathbb{N}^R,
\end{equation*}
where $\ell_i^{(r)}$ denotes the label assigned to $\bx_i$ by the base clustering method in projection $r$.\\
Two medoids $\bx_i$ and $\bx_j$ are assigned to the same consensus cluster if and only if their codes are identical:
\begin{equation*}
\Code(\bx_i) = \Code(\bx_j)
\,\Longleftrightarrow\,
\ell_i^{(r)} = \ell_j^{(r)} \quad \forall r \in \llbracket R \rrbracket.
\end{equation*}
In other words, a consensus cluster consists of medoids that co-occur in the same cluster across \emph{all} random feature views.
Any disagreement in a single view results in distinct codes.
The consensus construction therefore reduces to grouping medoids by identical codes, which can be implemented efficiently using hashing or sorting.
This encoding step is crucial, as it avoids explicit pairwise comparisons or graph constructions and enables the consensus operation to scale linearly with the number of samples.

This code-based formulation allows the consensus step to be implemented in near-linear time with respect to the number of samples, avoiding the quadratic cost of explicit pairwise comparisons. A detailed complexity analysis is provided in Appendix~\ref{compl:GetClusters}.

\paragraph{Output.}
The function \GetClusters{} outputs a label vector
$\Labels^{(e)} \in \llbracket n^{(e)} \rrbracket^{n^{(e-1)}}$,
where $n^{(e)}$ denotes the number of distinct codes.
Each entry of $\Labels^{(e)}$ specifies the consensus cluster assignment of an active medoid at iteration $e$.

\subsubsection{Medoid selection (\ChooseMedoids)}
\label{sec:ChooseMedoids}
At iteration $e$, the function \ChooseMedoids{} selects one representative per consensus cluster to define the next set of active medoids.
Let $\Labels^{(e)}$ denote the consensus labeling of the $n^{(e-1)}$ active medoids and let
$\mathcal{I}_k^{(e)} \subseteq \K^{(e-1)}$ be the index set of medoids assigned to consensus cluster $k \in \llbracket n^{(e)} \rrbracket$.

For each consensus cluster $k$, a single representative medoid is selected among the current active medoids in $\mathcal{I}_k^{(e)}$.
By default, we choose the medoid that is most central with respect to the other elements of the cluster, measured through a normalized inner-product criterion.
Specifically, all feature vectors are $\ell_2$-normalized and the selected medoid is
\begin{equation*}
j^* = \arg\min_{i \in \mathcal{I}_k^{(e)}}
\sum_{i' \in \mathcal{I}_k^{(e)}}
\left| \langle \tilde{\bx}_i^{(e)}, \tilde{\bx}_{i'}^{(e)} \rangle \right|,
\end{equation*}
where $\tilde{\bx} = \bx / \|\bx\|_2$.
This normalization is essential: without it, the inner product would be dominated by vector norms rather than angular similarity and the notion of centrality would be lost.

\begin{figure}[ht]
\centering
\includegraphics[width=0.8\linewidth]{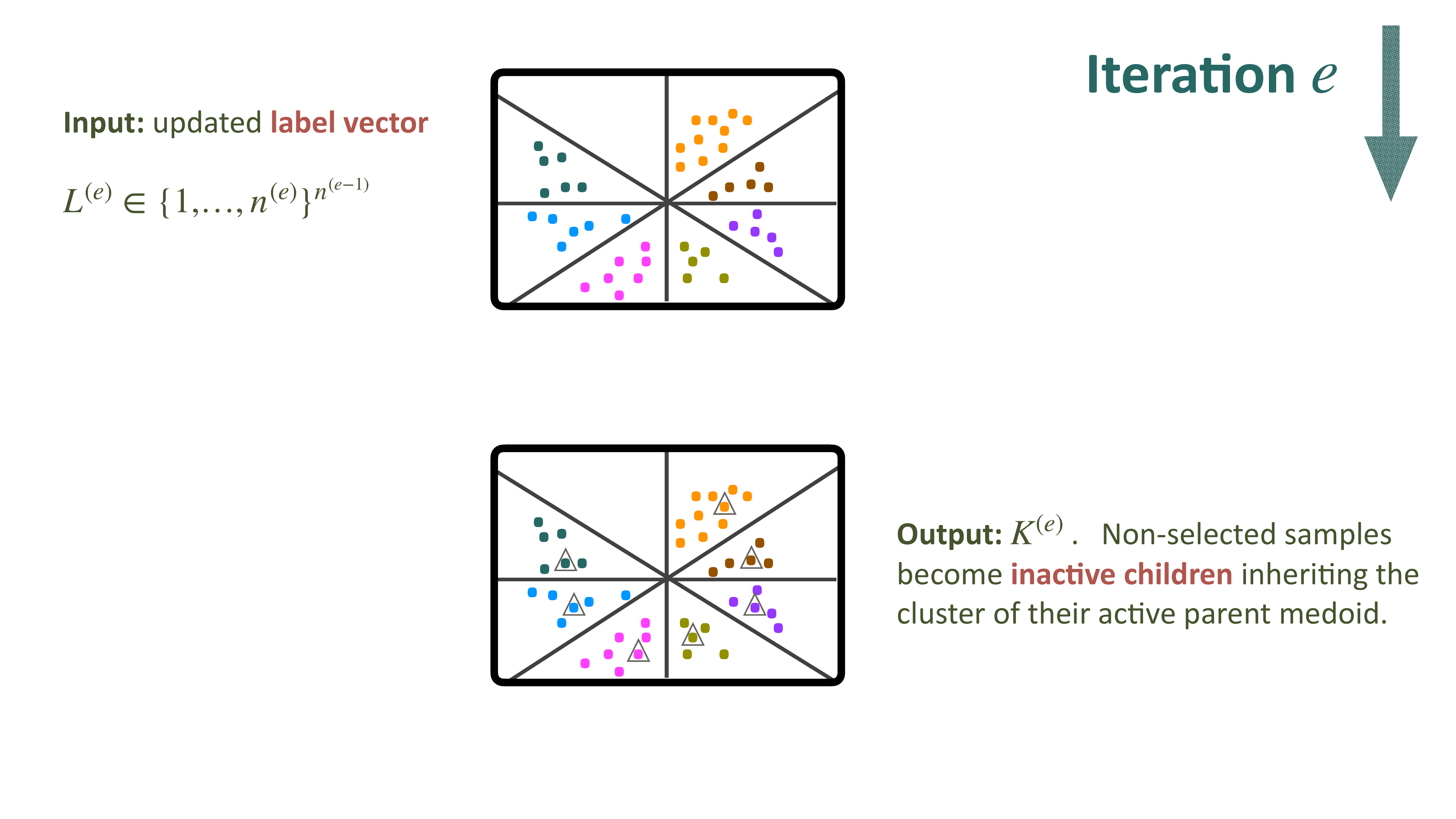}
\caption{\footnotesize Illustration of \ChooseMedoids{} at iteration $e$: one representative medoid is selected per consensus cluster; all other samples become inactive children.}
\label{fig:Step2}
\end{figure}

This criterion selects the sample that is most representative of the cluster in an angular sense.
While the normalized inner product is used \textbf{by default}, other similarity or distance measures can be employed depending on the base method and data geometry, for instance distances induced by an RBF kernel. Figure~\ref{fig:Step2} illustrates this step. 

A detailed analysis of the computational complexity of the medoid selection step, including the subsampling strategy used to ensure scalability, is provided in Appendix~\ref{compl:ChooseMedoids}.

\paragraph{Output.}
The function \ChooseMedoids{} outputs the updated index set $\K^{(e)}$ of active medoids.
All samples in $\mathcal{I}_k^{(e)}$ that are not selected become \textbf{inactive children}: they are removed from subsequent clustering steps but remain attached to their parent medoid for the construction of the Cluster Fusion Hierarchy (\CFH).

Using medoids rather than centroids provides additional interpretability,
since each representative corresponds to an actual data sample. Moreover, assigning new observations reduces to a nearest-medoid lookup, which is straightforward to implement and naturally compatible with the hierarchical structure produced by \CoHiRF{}.

To ensure scalability, medoid selection is performed using a subsampling
strategy within each consensus cluster, yielding a cost that is linear
in the number of clusters on average. A detailed complexity analysis is provided in Appendix~\ref{compl:ChooseMedoids}.

\subsubsection{Final labeling (\GetFinalLabels)}
\label{sec:get_final_labels}

The iterative procedure described above applies the functions \GetClusters{}, \ChooseMedoids{} and \UpdateParents{} until convergence.
At iteration $e$, the number of active medoids $n^{(e)}$ decreases as clusters merge. Convergence is reached either when no further merging occurs, i.e.,
$$
n^{(e)} = n^{(e-1)},
$$
which indicates that a stable partition of the data has been obtained, or when a maximum number of iterations is reached. At convergence, the remaining active medoids correspond to the roots of the hierarchical structure.

Once convergence is reached, the function \GetFinalLabels{} assigns a final cluster label to each original sample.
This is achieved by propagating labels from the active medoids at convergence down to all their descendants using the parent vector $\vP$. For each sample $i$, the final label is defined as the index of the root medoid reached by repeatedly following parent pointers:
$$
\Labels[i] = 
\begin{cases}
i & \text{if } \vP[i] = i \quad \text{(root: active medoid at convergence)},\\
\Labels[\vP[i]] & \text{otherwise} \quad \text{(recursively find root)}.
\end{cases}
$$
 This procedure assigns all samples that belong to the same fusion chain to a common final cluster.

The parent vector $\vP$ encodes the full hierarchy of successive cluster fusions.
Each sample corresponds to a leaf and each update of $\vP$ records a merge event between clusters across iterations.
We refer to the resulting hierarchical structure as the \textbf{Cluster Fusion Hierarchy (\CFH)}.
Importantly, the \CFH{} provides a multi-resolution representation of the data without requiring the selection of a cut level to obtain a final partition.

\paragraph{Output.}
The function \GetFinalLabels{} returns the final label vector $\Labels \in \llbracket n \rrbracket^{n}$ assigning each original sample to its final cluster.

The final labeling step runs in linear time and is negligible compared to the iterative clustering steps; a detailed analysis is provided in Appendix~\ref{compl:GetFinalLabels}.

\subsection{Relaxed consensus \RCoHiRF}
\label{sec:relaxed_consensus}

The strict consensus criterion retains only structures that are perfectly consistent across all random feature views.
While this provides strong robustness guarantees, it may become overly conservative in the presence of noisy or weakly informative projections.
In particular, some random feature subsets may be dominated by noise or non-discriminative variables, leading to partitions that strongly disagree with the majority of views.

The relaxed variant \RCoHiRF{} is designed to address this issue by selectively discarding such unreliable repetitions.
The underlying intuition is that repetitions which strongly disagree with the majority are likely to correspond to poor views and should not contribute to the consensus.

At iteration $e$, let $\Labels^{(e)}$ denote the consensus labeling obtained using the current set of repetitions.
For each repetition $r$, we compute a \textbf{leave-one-out agreement score}
$$
\ARI_{\mathrm{LOO}}(r)
=
\ARI\left(\Labels^{(e)},\, \Labels^{(e,-r)}\right),
$$
where $\Labels^{(e,-r)}$ denotes the consensus labeling obtained by excluding repetition $r$.
Here, $\Labels^{(e)}$ serves as a reference partition, while  $\ARI_{\mathrm{LOO}}(r)$ quantifies the impact of removing repetition $r$. 
We fix a threshold $h$ (set to $h = 0.8$ in our experiments) and apply the following iterative procedure.
First, the leave-one-out agreement scores are computed for all remaining repetitions.
Then, the repetition
$$
\widetilde r = \arg\min_r \ARI_{\mathrm{LOO}}(r)
$$
is identified as the most disagreeing one.
If $\ARI_{\mathrm{LOO}}(\widetilde r) < h$, this repetition is discarded and the procedure is repeated on the reduced set of repetitions.
The process stops when all remaining repetitions satisfy $\min_r \ARI_{\mathrm{LOO}}(r) \geq h$. This strategy removes only repetitions whose exclusion significantly alters the consensus, while preserving those that are broadly consistent with the majority. 


The relaxed consensus can be viewed as a stability-driven filtering mechanism. Rather than enforcing agreement across all views, \RCoHiRF{} retains a subset of mutually consistent projections, thereby reducing sensitivity to noisy or non-discriminative feature subsets.
In this context, the Adjusted Rand Index is not used as an external evaluation metric, but as an internal measure of consensus stability.
In particular, the relaxed consensus favors repetitions that preserve locally stable label relationships, while discarding views in which such relationships are disrupted by noise or non-informative features.

\paragraph{Output.}
The relaxed consensus is implemented as an option within the \GetClusters{} function.
Its output is a label vector
$
\Labels^{(e)} \in \llbracket n^{(e)} \rrbracket^{n^{(e-1)}},
$ 
where $n^{(e)}$ denotes the number of distinct consensus codes.
Each entry of $\Labels^{(e)}$ specifies the \emph{relaxed} consensus cluster assignment of an active medoid at iteration $e$.

The additional computational cost induced by the relaxed consensus is linear in the number of active medoids, up to a constant factor depending on the number of repetitions; a detailed analysis is provided in Appendix~\ref{compl:RelaxedConsensus}.

The effect of the relaxed consensus is illustrated in the robustness-to-noise experiment on the high-dimensional hypercube (Section~\ref{subsec:robustness_high_dim_noise_fixed_n}),
where we compare K-Means, \CoHiRF(K-Means) and its relaxed variant \RCoHiRF(K-Means). In this setting, \RCoHiRF{} mitigates the impact of spurious projections and improves robustness compared to strict consensus, while preserving the hierarchical structure induced by \CoHiRF{}.

\subsection{Batched extension for large-scale data}
\label{sec:batched}

To enable scalability to very large datasets and to operate under memory constraints, we introduce \CoHiRFb, a batched extension of \CoHiRF.
This extension leverages the hierarchical nature of the algorithm: medoids extracted from different subsets of the data can be progressively aggregated and merged without requiring access to the full dataset at once.

A batch size $b$ is selected based on hardware constraints and memory availability and the dataset is partitioned into batches of size $b$ (or as close as possible). Each batch is processed independently by applying a single iteration of \CoHiRF{}, yielding a set of representative medoids. The medoids obtained from all batches are then aggregated and repartitioned into new batches of comparable size. This procedure is repeated iteratively, forming successive batching levels in which the effective dataset size decreases as medoids are merged. When only a single batch remains, \CoHiRF{} is applied until full convergence.

To prevent abrupt variations in batch composition across levels, one randomly selected batch is temporarily held aside at each level, ensuring that the number of samples per batch remains approximately stable. Empirically, this stabilization improves the aggregation of medoids originating from different batches and preserves compatibility with fixed hyperparameters across levels.

\paragraph{Scalability and practical considerations.} The batched formulation allows batches to be processed independently and in parallel. As a result, memory usage is reduced from that required to process all $n$ samples simultaneously to that required for a single batch of size $b$. Assuming a geometric reduction in the number of medoids across batching levels, the number of levels grows logarithmically with $n$, enabling near-linear scaling in practice.

Smaller batch sizes reduce memory requirements but may slightly fragment global structure when the base method relies on weak or unstable label assignments, whereas larger batches better preserve global consistency at the cost of increased memory usage. While batching introduces additional overhead due to projection, consensus and hierarchical aggregation, it extends the applicability of \CoHiRF{} well beyond the memory limits of standard clustering algorithms. Formal time and memory complexity guarantees for the batched formulation are given in Theorems~\ref{thm:batched_complexity} and~\ref{thm:space_complexity}.

\section{Overall complexity study}
\label{sec:overall_complexity}

We now synthesize the complexity results of the different components of \CoHiRF{} into an overall analysis of its computational and memory requirements. All detailed derivations and intermediate bounds are provided in Appendix~\ref{app:complexity_proofs}.

\subsection{Computational complexity}

We first analyze the computational complexity of the standard (non-batched)
formulation of \CoHiRF{}. The algorithm proceeds iteratively by applying a base clustering method to random feature projections of a progressively shrinking set of active medoids.

\begin{theorem}[Overall computational complexity of \CoHiRF{}]
\label{thm:complexity}
Let $n$ denote the number of samples, $p$ the number of features, $R$ the number of random feature projections and $q$ the projection dimension.
Let $\TimeComplexity{\BCM{}, m, q}$ denote the time required by the base clustering method (\BCM{}) to process $m$ samples in dimension $q$.

Then, the total computational complexity of \CoHiRF{} is
$$
\TimeComplexity{\CoHiRF}
= 
O\left(R \,\sum_{e=1}^{E}\TimeComplexity{\BCM{}, n^{(e-1)}, q}\right),
$$
where $n^{(e-1)}$ denotes the number of active medoids at iteration $e$ and $E$ is the number of iterations until convergence.

If the number of active medoids decreases geometrically across iterations, i.e., $n^{(e)} \leq \alpha \cdot n^{(e-1)} \quad \text{for some } \alpha < 1,$ and if $\TimeComplexity{\BCM{}, m, q}$ is linear in $m$, then the overall complexity reduces to
$$
\TimeComplexity{\CoHiRF}
= O\left(R \cdot \TimeComplexity{\BCM{}, n, q}\right)\quad\text{(on average)}.
$$
In contrast, in the slow-convergence regime (worst case), where only a constant number of medoids merge at each iteration, the number of iterations satisfies $E = O(n)$ and the total complexity becomes quadratic in $n$.
\end{theorem}

The proof of Theorem~\ref{thm:complexity} is given in Appendix~\ref{app:proof:thm:complexity}.

When K-Means is used as the base clustering method with $k$ clusters and $I$ iterations per run, we have
$
\TimeComplexity{\text{K-Means}, m, q} = O(m \, q \, k \, I).
$
Under the geometric reduction assumption of Theorem~\ref{thm:complexity},
this yields 
$
\TimeComplexity{\CoHiRF}
= O(R \, n \, q \, k \, I),
$
corresponding to running K-Means $R$ times on low-dimensional random projections.


We now consider the batched extension \CoHiRFb{}, designed to enable
parallel execution and large-scale processing.

\begin{theorem}[Time complexity of the batched formulation]
\label{thm:batched_complexity}
Let $n$ denote the total number of samples, $b$ the batch size and
$B = \lceil n / b \rceil$ the number of batches.
Let $\TimeComplexity{\BCM{}, m, q}$ denote the time required by the base clustering
method to process $m$ samples in dimension $q$. In the batched formulation \CoHiRFb{}, the time complexity satisfies:

\medskip
\noindent
\textbf{(i)} 
 When batches are processed \textbf{sequentially}, the total runtime is
$$
\TimeComplexity{\text{batched, seq}}=O\left(B \, R \cdot \TimeComplexity{\BCM{}, b, q}\right).
$$
Moreover, if the base clustering method has linear complexity in the number of samples,
i.e.\ $\TimeComplexity{\BCM{}, b, q} = O(b\, g(q))$,
this simplifies to
$$
\TimeComplexity{\text{batched, seq}}=O\left(R \, n\, g(q)\right),
$$
since $B = \lceil n / b \rceil$, matching the average-case complexity of the non-batched formulation.

\medskip
\noindent
\textbf{(ii)}  When batches are processed \textbf{in parallel} and assuming a geometric reduction
in the number of medoids across batching levels, the parallel time complexity is
$$
\TimeComplexity{\text{batched, parallel}}=O\left(\log_b(n) \cdot R \cdot \TimeComplexity{\BCM{}, b, q}\right).
$$

For typical batch sizes, $\log_b(n)$ remains small, so the parallel time complexity
depends only weakly on the total dataset size.
\end{theorem}

The proof is provided in Appendix~\ref{app:proof:thm:batched_complexity}.

\subsection{Memory complexity}

\medskip
Finally, we analyze the memory requirements of \CoHiRF{}.

\begin{theorem}[Space complexity of \CoHiRF{}]
\label{thm:space_complexity}
Let $n$ denote the number of samples, $p$ the original feature dimension, $q$ the projection dimension and $R$ the number of random feature views.
Let $\text{Memory}(\BCM{}, m, q)$ denote the memory required by the base clustering method to process $m$ samples in dimension $q$.

\textbf{(i)}  In the \textbf{non-batched formulation}, the peak memory usage of \CoHiRF{} is
$$
O\left(n \, p + R \, n \, q + \text{Memory}(\BCM{}, n, q)\right),
$$
where the first term corresponds to storing the original data, the second to the projected views and consensus codes and the third to the base clustering method.


\textbf{(ii)}  In the \textbf{ batched formulation} with batch size $b$, the peak memory usage is reduced to
$$
O\left(b \, p + R \, b \, q + \text{Memory}(\BCM{}, b, q)\right),
$$
independently of the total number of samples $n$.


\end{theorem}

The proof of Theorem~\ref{thm:space_complexity} is given in
Appendix~\ref{app:proof:thm:space_complexity}.

\section{Experiments}
\label{sec:experiments}

The \CoHiRF{} framework is implemented in Python using NumPy. Base clustering methods are implemented using \texttt{scikit-learn} \citep{pedregosa2011} or reimplemented in python based on their original work implementation. All experiments were conducted on a machine equipped with an Intel(R) Xeon(R) CPU E5-2630 v4 processor and up to 126~GB of RAM. The source code used to run the experiments is available at our GitHub repository \hyperlink{https://github.com/BrunoBelucci/cohirf/settings}{https://github.com/BrunoBelucci/cohirf/settings}.

\subsection{Experimental protocol}

Our experimental study is designed to understand how CoHiRF behaves across controlled and real-world scenarios, rather than to provide an exhaustive benchmark. Synthetic datasets are used to isolate specific mechanisms, while real-world datasets illustrate how these mechanisms interact with practical data structures.

Note that \CoHiRF{} introduces an additional computational cost proportional to the number of feature views. In practice, this overhead remains moderate and is largely offset by parallelization and batching, while enabling clustering methods that would otherwise be infeasible at scale (see Appendix~\ref{app:practical}).

\subsubsection{Evaluation metrics and hyperparameter tuning}
\label{subsec:metrics_hyperpara}

A common definition of a cluster describes it as a set of elements that are internally similar and externally dissimilar.  This definition, however, is incomplete without specifying the notion of similarity, which is implicitly determined by the chosen clustering objective and representation.

To evaluate clustering quality, we rely on three complementary criteria capturing distinct aspects of clustering performance. \textbf{The Adjusted Rand Index (\ARI)} \citep{ari1985} measures agreement between predicted clusters and reference labels, providing a supervised criterion of external validity. It answers the question of whether the inferred partition is consistent with a given ground truth.
\textbf{The Silhouette score (\Sil)} \citep{silhouette1987} is an unsupervised internal validation metric that evaluates intra-cluster cohesion and inter-cluster separation. It quantifies whether each point is, on average, closer to points within its own cluster than to points in other clusters.
Finally, \textbf{the Calinski--Harabasz index (\CH)} \citep{calinsky1974} measures the ratio of between-cluster dispersion to within-cluster variance, emphasizing the compactness and separation of the resulting partition.

\subsubsection{Computational considerations and hyperparameter optimization}
\label{subsec:computational-consideration}

Each clustering algorithm is evaluated using tuned hyperparameters selected to yield the best achievable performance under the considered criterion.
Unless otherwise specified, we employ 60 optimization trials of the Tree-structured Parzen Estimator (TPE) \citep{tpe2023} for hyperparameter search.
The hyperparameter search space for \CoHiRF{} and variants\footnote{All CoHiRF variants use the same hyperparameter search space as their base clustering methods, in addition to the listed CoHiRF-specific parameters. For CoHiRF(SC-SRGF), feature subsampling is disabled ($q=p$) and the number of repetitions $R$ is fixed (to $R=1$ or $R=2$ depending on the experiment) and not tuned.The relaxed consensus threshold $h$ applies only to relaxed variants and is not used with SC-SRGF.} is reported in Table~\ref{tab:ourmethod_hyperparameter_search_spaces}, while the search spaces for the base clustering methods are detailed in Appendix~\ref{app:algo-search-space}.

For each dataset and each clustering method, hyperparameters are tuned \emph{independently for each evaluation metric}. Concretely, for every method we perform three separate hyperparameter optimization procedures: (i) \textbf{\ARI-tuned}, where hyperparameters are selected to maximize the Adjusted Rand Index when reference labels are available, (ii) \textbf{\CH-tuned}, where hyperparameters are selected to maximize the Calinski--Harabasz index and (iii) \textbf{\Sil-tuned}, where hyperparameters are selected to maximize the Silhouette score.

As a result, the value reported in a given column of the tables corresponds to the performance obtained under the tuning protocol specific to that metric. Each reported score is averaged over multiple independent runs with different random seeds and we report the mean together with the standard deviation. All confidence intervals displayed in the figures correspond to 95\% intervals, computed as percentile intervals from the bootstrap distribution obtained over at least five independent runs.

\begin{table}[ht]
\caption{Hyperparameter search spaces for \CoHiRF{} used in our experiments.}
\label{tab:ourmethod_hyperparameter_search_spaces}
\begin{center}
\begin{tabular}{lll}
\toprule
Model & Parameter & Values \\
\midrule
\multirow{4}{*}{CoHiRF and variants} 
 & Percentage of features ($q_{\%}$) & Float[0.1, 1.0] \\
 & Repetitions ($R$) & Int[2, 10] \\
 & Relaxed threshold ($h$) & $h = 0.8$ \\
 & Number of batches ($B$) & $10$ (default), $100$ \\
\bottomrule
\end{tabular}
\end{center}
\end{table}

This evaluation protocol is intentionally metric-specific.
For each (dataset, method, metric) triplet, hyperparameters are optimized
independently with respect to the corresponding evaluation criterion.
Reported values therefore correspond to metric-specific optimal configurations,
rather than to a single set of hyperparameters evaluated across all metrics.
In particular, internal validation metrics are not evaluated using
hyperparameters optimized for external agreement and vice versa.
This protocol avoids conflating algorithmic performance with mismatched
hyperparameter choices.



\subsection{Experimental study: synthetic analysis}
\label{sec:experiments_synthetic}

\subsubsection{Interpretable hierarchies and adaptive cluster selection (Iris)}
\label{subsec:toy_iris}

Beyond clustering accuracy, \CoHiRF{} produces an explicit hierarchical structure that can be inspected without access to ground-truth labels. This property is particularly relevant in unsupervised settings, where external validation criteria are unavailable and model selection must rely on structural information derived from the data.

We illustrate this behavior on the Iris dataset, which contains $n=150$ samples described by $p=4$ features and is commonly associated with $C=3$ botanical species. We consider K-Means and a kernel K-Means variant (\KK{}), together with their \CoHiRF{} counterparts. \textbf{Kernel K-Means} \citep{dhillon2004} is implemented using an RBF kernel, approximated via random Fourier features \citep{rahimi2007}.  Unless stated otherwise, we use a fixed approximation with 500 random Fourier features throughout all experiments.



Clustering results are reported in Table~\ref{tab:iris}. For both K-Means and \KK{}, applying \CoHiRF{} is associated with higher Adjusted Rand Index (ARI) values. The relaxed consensus variant achieves the highest ARI in both cases. Internal validation metrics (Calinski--Harabasz and Silhouette scores) remain comparable to those of the base methods, indicating that the improvement in external agreement is not accompanied by a degradation of internal cluster structure.

\begin{table}[ht]
\centering
\caption{Clustering results on Iris dataset, $n=150$, $p=5$, $C=3$.}
\label{tab:iris}
\begin{tabular}{lllll}
\toprule
Model & \ARI & \CH & \Sil  \\
\midrule
\color{blue}{K-Means} & 0.618 $\pm$ 0.035 & \bfseries 241.038 $\pm$ 0.000 & \bfseries 0.579 $\pm$ 0.000 \\
CoHiRF & 0.786 $\pm$ 0.132 & \underline{238.928 $\pm$ 4.719} & \bfseries 0.579 $\pm$ 0.000   \\
\RCoHiRF & \underline{0.868 $\pm$ 0.079} & \bfseries 241.038 $\pm$ 0.000 & \bfseries 0.579 $\pm$ 0.000  \\
\color{blue}{KernelRBFK-Means (\KK)} & 0.641 $\pm$ 0.022 & 228.040 $\pm$ 29.064 & \underline{0.578 $\pm$ 0.002} \\
CoHiRF(\KK) & 0.798 $\pm$ 0.094 & 231.460 $\pm$ 21.417 & \bfseries 0.579 $\pm$ 0.000  \\
\RCoHiRF(\KK) & \bfseries 0.900 $\pm$ 0.018 & 233.757 $\pm$ 16.280 & \bfseries 0.579 $\pm$ 0.000  \\
\bottomrule
\end{tabular}
\end{table}

\begin{figure}[ht]
    \centering
    \includegraphics[width=0.6\linewidth]{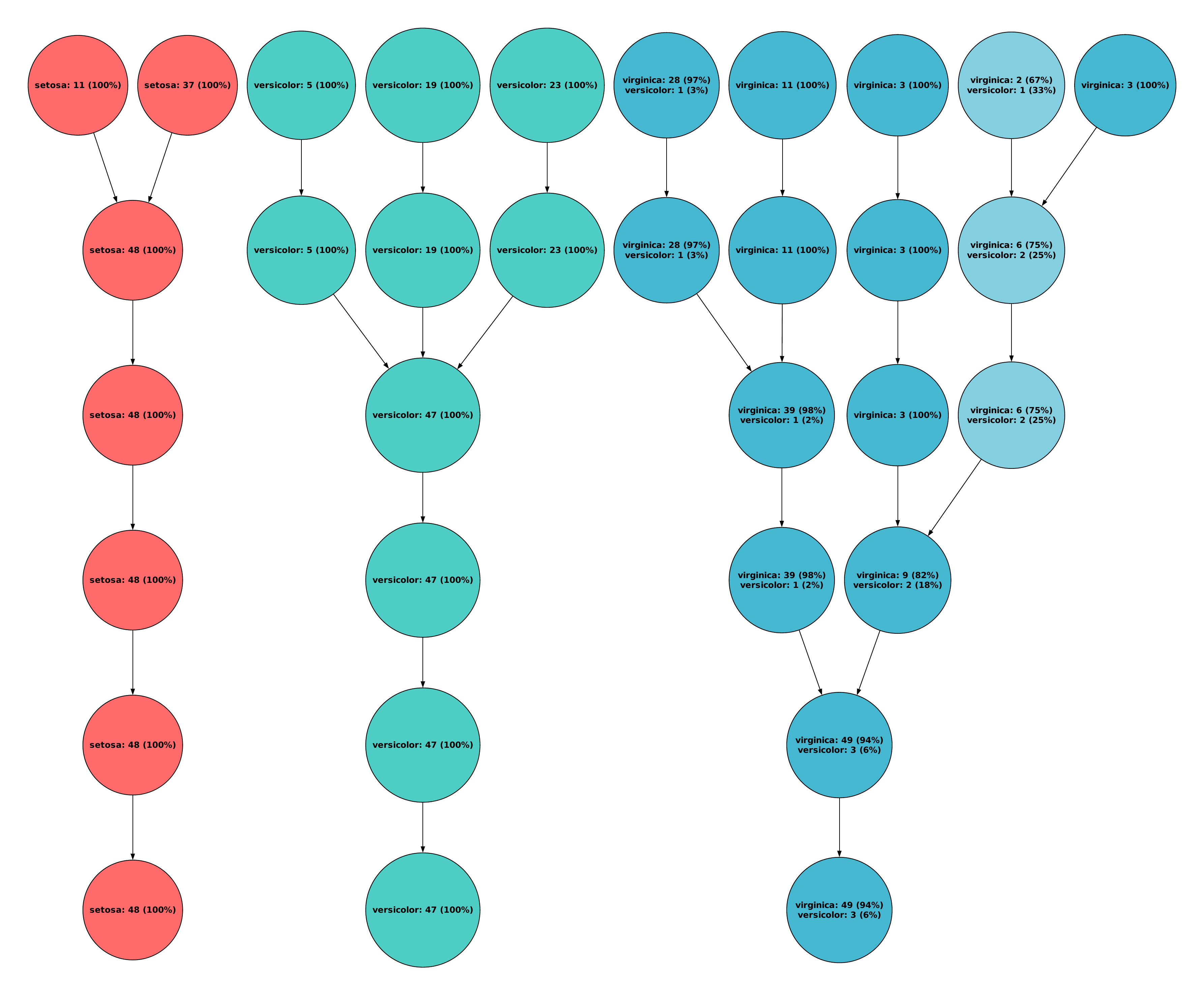}
    \caption{\CFH{} obtained with \RCoHiRF(\KK) on the Iris dataset.}
    \label{fig:dendogram-iris}
\end{figure}

Beyond these quantitative results, \CoHiRF{} produces an explicit Cluster Fusion Hierarchy (\CFH) that records how groups of samples merge across iterations. Figure~\ref{fig:dendogram-iris} displays the hierarchy obtained with \RCoHiRF(\KK). The hierarchy shows that samples corresponding to the \textit{setosa} species form a stable cluster early in the process, while samples from \textit{versicolor} and \textit{virginica} merge more gradually. This progressive fusion is consistent with the known overlap between these two species in the Iris feature space.

The \CFH{} provides direct access to the relative ordering and scale at which clusters merge. Early merges correspond to label relations that are consistently preserved across iterations, while later merges indicate boundaries that are less stable. Unlike classical agglomerative clustering, \CoHiRF{} is not restricted to pairwise merges: all samples sharing identical label codes across views are contracted simultaneously into a single representative. As a result, the hierarchy captures consensus-driven structure at multiple resolutions and is available independently of external ground-truth labels.

\subsubsection{Robustness to high-dimensional noise (fixed \texorpdfstring{$n$}{n})}
\label{subsec:robustness_high_dim_noise_fixed_n}

A central motivation behind \CoHiRF{} is to improve the reliability of clustering methods in high-dimensional settings. Even when the underlying cluster structure is simple and well separated in a low-dimensional subspace, the presence of many non-informative coordinates can severely degrade distance-based methods due to distance concentration. As dimensionality increases, informative signal is progressively overwhelmed by noise, making single-view clusterings unstable and sensitive to spurious variations.

\paragraph{Synthetic hypercube model (fixed $n=10^3$, increasing noise).}
We consider a controlled synthetic setting where the intrinsic cluster structure is fixed while the  dimension increases. We generate a mixture of $C=5$ Gaussian clusters in an informative subspace of dimension $p_{\inf}=3$, with $n=1000$ samples in total. Cluster centers are sampled from the vertices of a $p_{\inf}$-dimensional hypercube and observations are drawn as
$
\bx \mid c \sim \mathcal{N}(\mu_c,\, I_{p_{\inf}}),
$
with identical within-cluster dispersion. The hypercube edge length is set to $a = 6\sqrt{p_{\inf}}$, ensuring a stable separation-to-noise ratio in the informative subspace. We then append $p_{\mathrm{noise}}$ independent Gaussian noise dimensions,
$
\bx = (\bx_{\inf},\, \bx_{\mathrm{noise}}), \qquad
\bx_{\mathrm{noise}} \sim \mathcal{N}(\boldsymbol{0}_{p_{\mathrm{noise}}}, I_{p_{\mathrm{noise}}}),
$
and progressively increase $p_{\mathrm{noise}}$ from $10$ to $10^{4}$, while keeping $n$ and the informative structure fixed. This construction isolates the effect of high-dimensional noise without confounding it with changes in sample size or intrinsic geometry.

We compare standard K-Means with \CoHiRF(K-Means) and its relaxed variant \RCoHiRF(K-Means). The relaxed version replaces strict consensus by the leave-one-out filtering strategy described in Section~\ref{sec:relaxed_consensus}, designed to discard unstable repetitions arising from poorly informative feature views.

\begin{figure}[ht]
\centering
\includegraphics[width=0.8\linewidth]{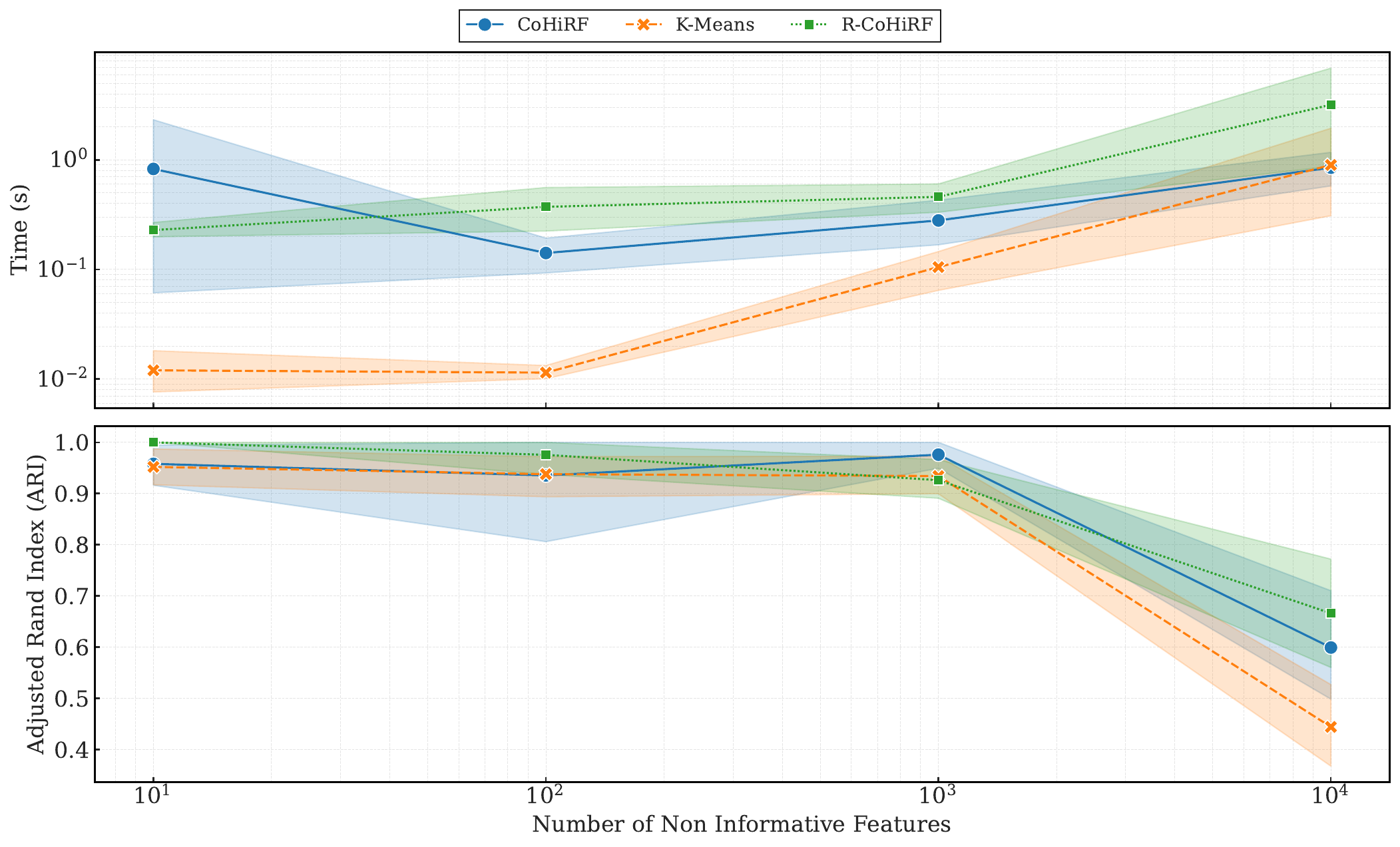}
\caption{Adjusted Rand Index (\ARI) and runtime for K-Means, \CoHiRF(K-Means) and \RCoHiRF(K-Means) on hypercube mixtures, as the number of non-informative features increases from $p_{\mathrm{noise}}=10$ to $p_{\mathrm{noise}}=10^{4}$ (with fixed $n=1000$ and $C=5$).}
\label{fig:hypercube-non-informative}
\end{figure}

Figure~\ref{fig:hypercube-non-informative} reports the \ARI{} as a function of $p_{\mathrm{noise}}$. As expected, all methods degrade as noise dimensions dominate. However, both \CoHiRF(K-Means) and \RCoHiRF(K-Means) substantially mitigate this degradation, retaining significantly higher \ARI{} values than standard K-Means in high-dimensional regimes. Notably, meaningful partitions remain detectable even for $p_{\mathrm{noise}} = 10^{4}$.

The relaxed variant consistently provides the strongest robustness. This behavior is consistent with the idea that, in very noisy settings, some random feature views are highly misleading. By discarding repetitions that disrupt locally stable label relationships, relaxed consensus preserves cluster structure that is stable across most projections, while filtering noise-driven artifacts.

Compared with a single run of K-Means, \CoHiRF{} incurs additional computational cost due to repeated clusterings and consensus aggregation. In this experiment, the overhead remains moderate, since each repetition operates on low-dimensional projections. This additional cost is outweighed by the robustness gains observed in the most challenging high-dimensional regimes, where standard K-Means rapidly becomes unreliable.

\subsubsection{Scalability in \texorpdfstring{$n$}{n} under high-dimensional noise}
\label{subsec:scalability_n_high_dim_noise}

We now move to a large-scale regime and study how \CoHiRF{} enables clustering methods beyond K-Means to remain practical when both the feature dimension and the sample size are large.
This experiment focuses on spectral clustering and highlights the role of batching when memory constraints become critical.

\paragraph{Synthetic hypercube model (fixed $p=10^4$, increasing $n$).}
We generate a Gaussian mixture with $C=5$ clusters in an informative subspace of dimension $p_{\inf}=10$. Cluster centers are sampled from the vertices of a $p_{\inf}$-dimensional hypercube and observations are drawn as 
$
\bx \mid c \sim \mathcal{N}(\mu_c,\, I_{p_{\inf}}),
$
with controlled within-cluster dispersion. The hypercube edge length is set to $a = 6\sqrt{p_{\inf}}$, ensuring consistent separation in the informative subspace. We then append $p_{\mathrm{noise}}=10^4$ independent Gaussian noise dimensions and vary the sample size from $n=10^2$ to $n=10^5$, keeping the dimension fixed. Compared with Section~\ref{subsec:robustness_high_dim_noise_fixed_n}, the higher intrinsic dimension makes the signal stronger, allowing K-Means to remain a competitive baseline despite the presence of many non-informative features. Graph-based methods can nevertheless provide additional benefits by exploiting local neighborhood structure.

We compare (i) K-Means, (ii) SC-SRGF, a scalable spectral clustering method and (iii) \CoHiRFb(SC-SRGF), the batched variant of \CoHiRF{}(SC-SRGF). For this experiment, we disable feature subsampling ($q=p$) and fix $R=1$, since randomness arises from the graph construction rather than from multiple feature views.

Here, the role of \CoHiRF{} is not to improve clustering accuracy through feature perturbations, but to make spectral clustering feasible at large scale. This is achieved by progressively reducing the problem size through medoid-based compression, while preserving the full hierarchical structure. Importantly, this procedure is not bagging: clustering decisions are propagated through the hierarchy rather than averaged across independent runs.
\begin{figure}[ht]
    \centering
    \includegraphics[width=0.8\linewidth]{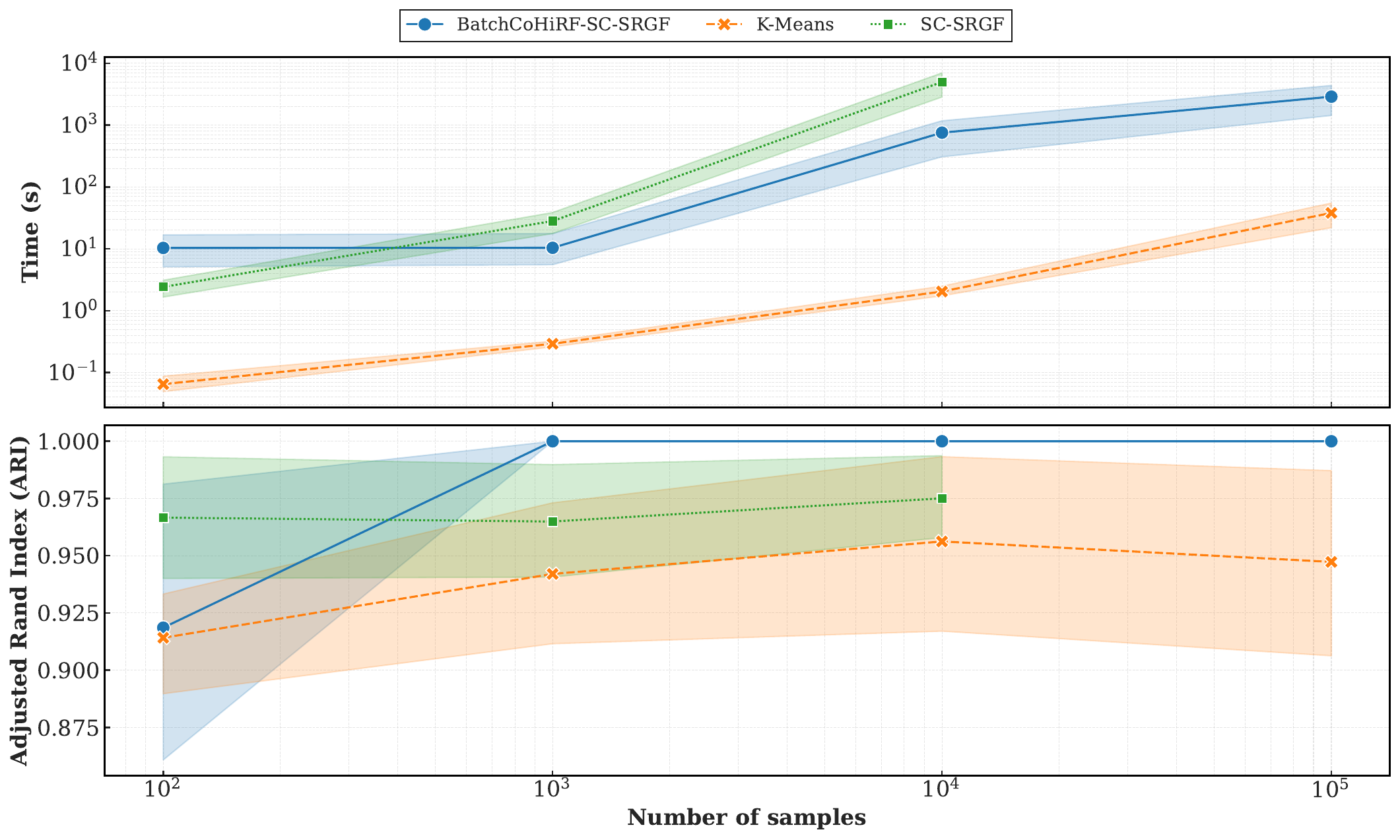}
\caption{Adjusted Rand Index (ARI) and runtime for K-Means, SC-SRGF and \CoHiRFb(SC-SRGF) on hypercube mixtures, with fixed $p_{\inf}=10$ and $p_{\mathrm{noise}}=10^4$, as the sample size increases from $n=10^2$ to $n=10^5$.}
    \label{fig:hypercube-samples}
\end{figure}

Figure~\ref{fig:hypercube-samples} reports \ARI{} and runtime as a function of $n$.
K-Means performs well in this setting, with high \ARI{} values across all scales, but exhibits larger variability across runs. SC-SRGF consistently improves over K-Means in both accuracy and stability as long as it can be executed.

The batched variant \CoHiRFb(SC-SRGF) further improves stability.
Starting at $n=10^3$, it achieves perfect clustering ($\ARI=1$) with virtually no variability and maintains this performance up to $n=10^5$. This indicates that hierarchical medoid-based reduction not only improves scalability, but also acts as a strong stabilizing mechanism by suppressing noise-induced graph artifacts in this experiment.

K-Means is the fastest method at all scales. SC-SRGF is substantially more expensive and becomes infeasible beyond moderate sample sizes due to our memory constraints. When SC-SRGF is feasible, \CoHiRFb(SC-SRGF) exhibits two regimes: it can be slightly slower at small scales due to batching overhead, but becomes consistently faster once graph construction dominates the cost.
Our goal is not to outperform the base method in raw runtime, but to extend its applicability.
In this sense, \CoHiRFb(SC-SRGF) enables spectral clustering to operate in regimes where SC-SRGF cannot run at all, while preserving perfect clustering quality and stability.

\subsubsection{Non-convex structure and large-\texorpdfstring{$n$}{n} scalability}
\label{subsec:nonconvex_large_n}

We now consider a setting with a strongly non-convex cluster structure. This experiment illustrates how \CoHiRF{} can be used to extend the applicability of a density-based clustering method to very large sample sizes, in a setting where centroid-based approaches are not appropriate.

\begin{figure}[ht]
    \centering
    \begin{subfigure}[b]{0.32\textwidth}
        \centering
        \includegraphics[width=\textwidth]{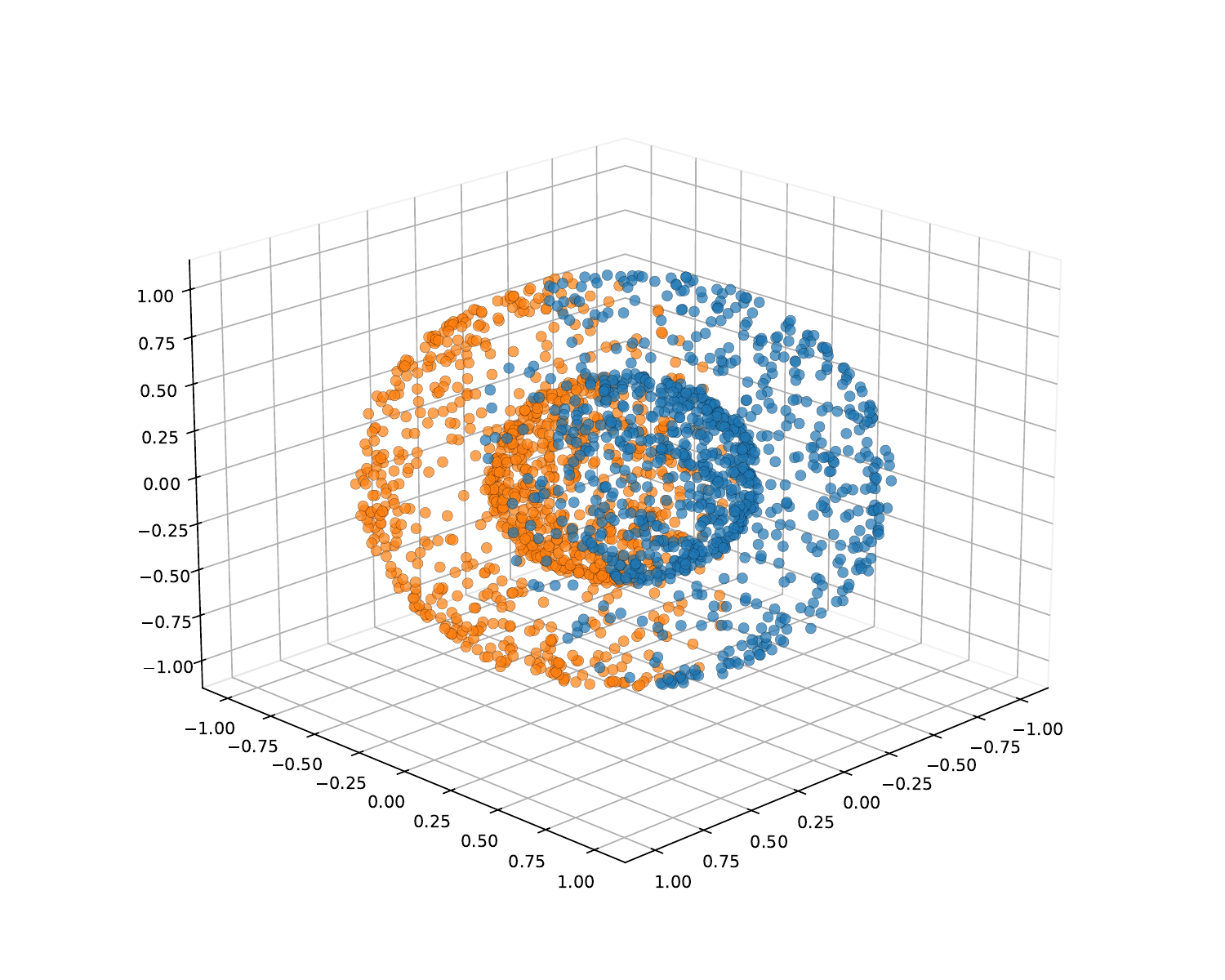}
        \caption{K-Means (2 clusters)}
        \label{fig:spheres-3d-results-K-Means-2}
    \end{subfigure}
    \hfill
    \begin{subfigure}[b]{0.32\textwidth}
        \centering
        \includegraphics[width=\textwidth]{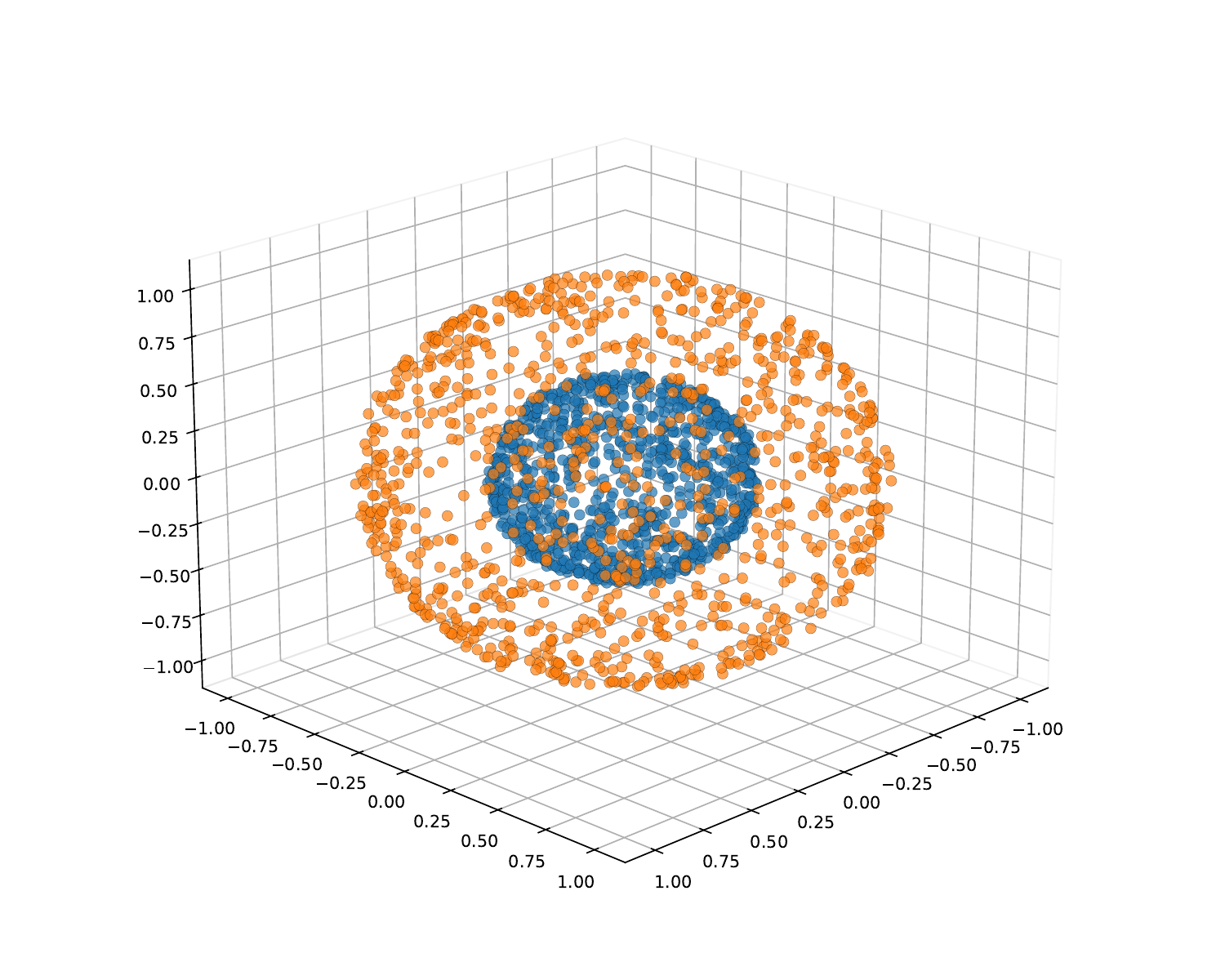}
        \caption{DBSCAN}
        \label{fig:spheres-3d-results-dbscan}
    \end{subfigure}
    \hfill
    \begin{subfigure}[b]{0.32\textwidth}
        \centering
         \includegraphics[width=\textwidth]{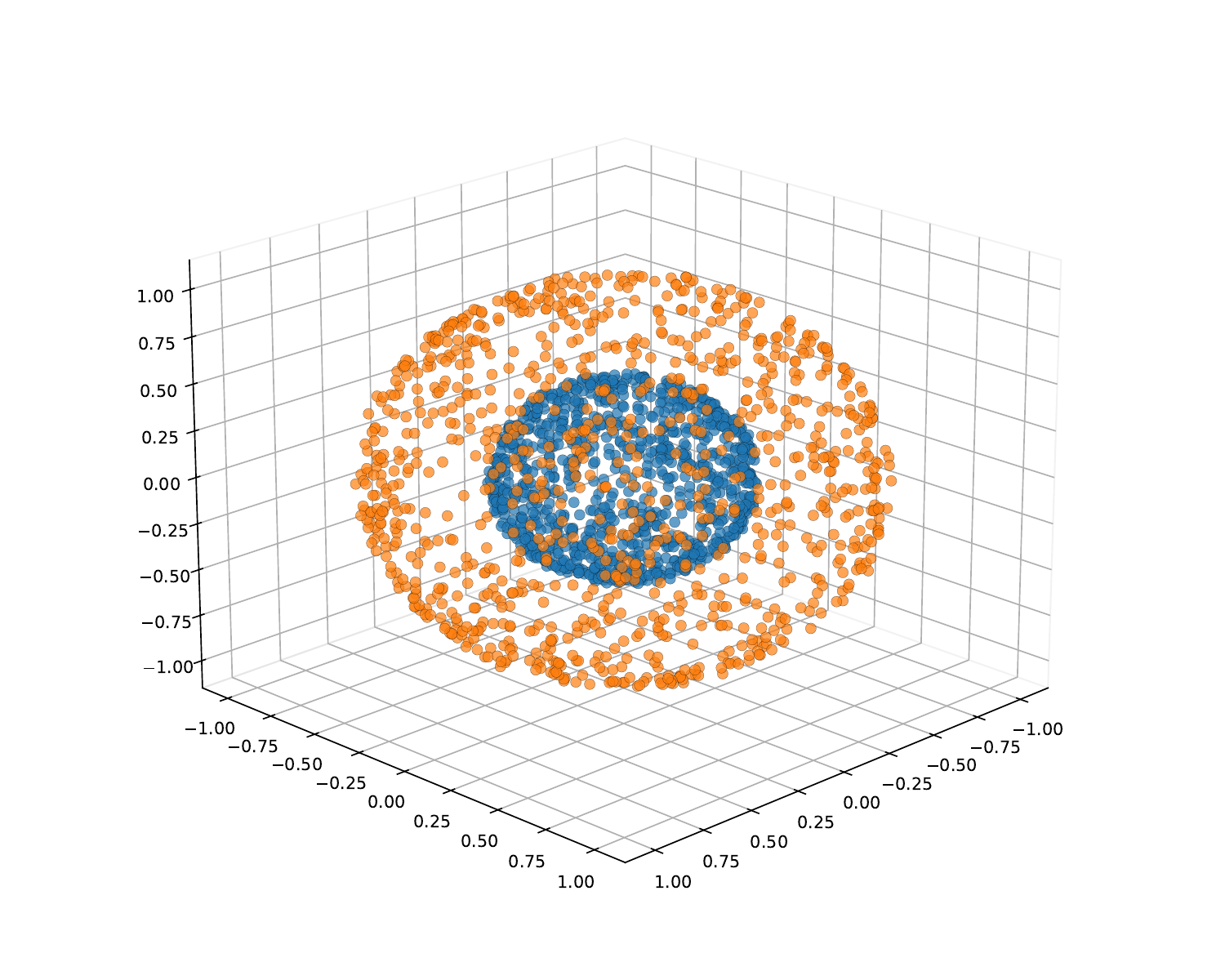}
        \caption{CoHiRF(DBSCAN)}
        \label{fig:spheres-3d-results-cohirf-dbscan}
    \end{subfigure}
    \caption{Clustering results on two concentric spheres with radii $0.5$ and $1.0$ (radial noise standard deviation $0.01$), using $2\times 1000$ samples.}
    \label{fig:spheres-3d-results-images}
\end{figure}

\paragraph{Concentric spheres model.}
We generate a two-cluster dataset by sampling points on the surfaces of two concentric spheres in $\mathbb{R}^3$, with radii $0.5$ and $1.0$. Each point is perturbed by radial Gaussian noise with standard deviation $0.01$. We first consider a visualization setting with $1000$ samples per sphere (Figure~\ref{fig:spheres-3d-results-images}) and then scale the same generative process up to $n=2\times10^7$ samples.


Because both clusters share the same centroid in expectation, centroid-based methods such as K-Means are not suited to this geometry and fail even when the true number of clusters is provided  (Figure~\ref{fig:spheres-3d-results-images}-(a)). In contrast, this structure is well matched to density-based clustering: the two clusters correspond to dense surfaces separated by an empty region, which aligns directly with the density-connectivity principle used by DBSCAN. When it can be executed, DBSCAN correctly recovers the two clusters (Figure~\ref{fig:spheres-3d-results-images}-(b)). However, standard DBSCAN becomes infeasible beyond moderate sample sizes under our memory constraints, typically around $n \approx 2\times10^4$. To address this limitation, we apply the batched variant \CoHiRFb(DBSCAN). The dataset is partitioned into batches of fixed size, DBSCAN is applied within each batch and representative medoids are aggregated hierarchically across iterations. Batch sizes are chosen so that DBSCAN remains feasible within each batch.

%


%
\begin{figure}[ht]
    \centering
    \includegraphics[width=0.6\linewidth]{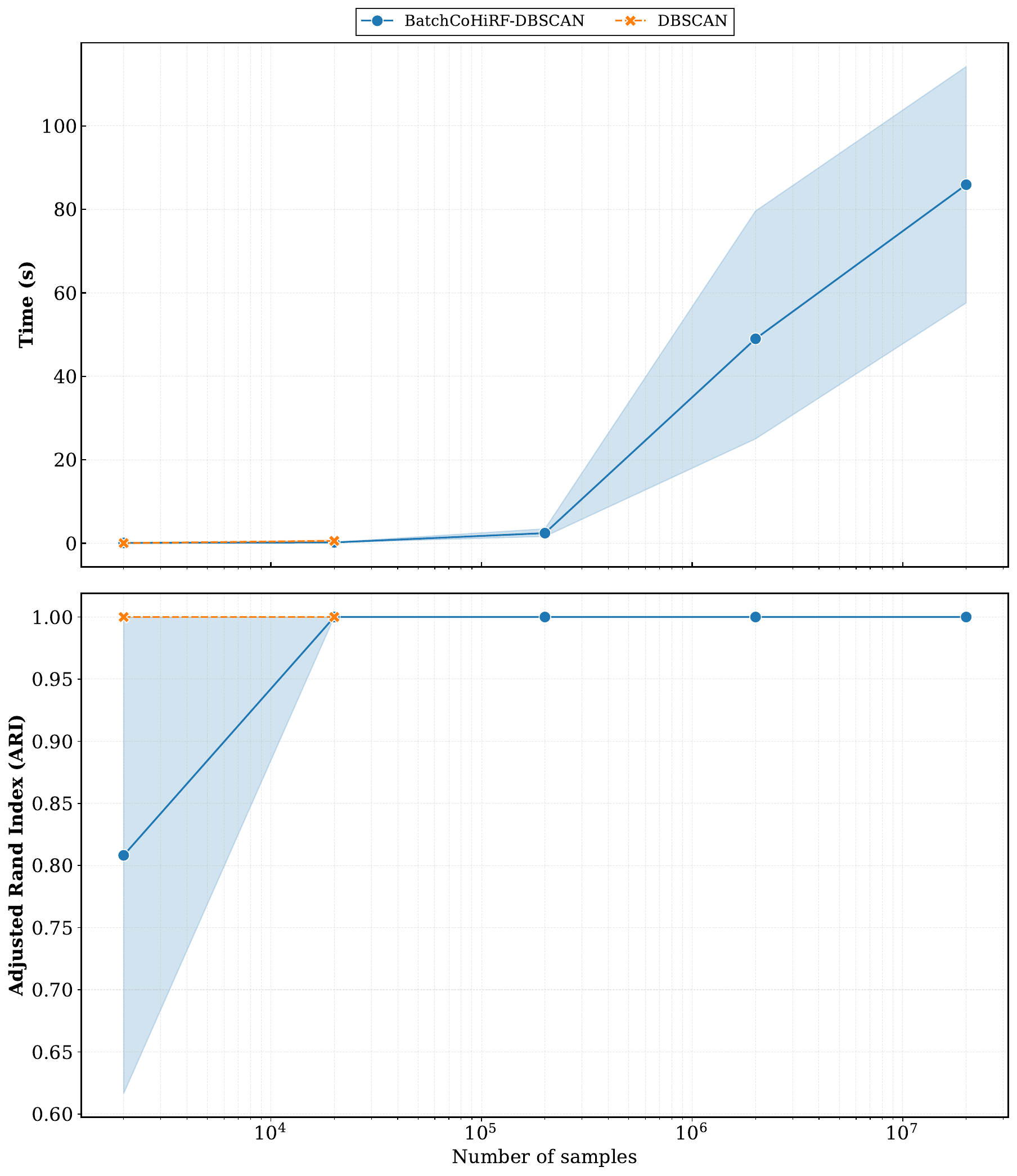}
    \caption{\ARI{} for DBSCAN and \CoHiRFb(DBSCAN) on concentric spheres, obtained while increasing the number of samples from $2\times 10^{3}$ to $2\times 10^{7}$. Standard DBSCAN becomes infeasible beyond $2\times10^{4}$ under our memory constraints.}
    \label{fig:spheres-ari}
\end{figure}
%

\newpage
Figure~\ref{fig:spheres-ari} reports the \ARI{} as a function of the sample size. When feasible, DBSCAN achieves perfect clustering ($\ARI=1$). At small scales, \CoHiRFb(DBSCAN) slightly underperforms due to batching overhead. As the sample size increases, standard DBSCAN becomes infeasible, while \CoHiRFb(DBSCAN) continues to scale and maintains perfect clustering up to $n=2\times10^7$. Consistently with Theorem~\ref{thm:batched_complexity}, the runtime of \CoHiRFb(DBSCAN) is governed by the cost of applying the base method to batches rather than to the full dataset. While batching introduces a small overhead at the smallest scales, it enables density-based clustering in regimes where the original method cannot be applied.

%


\subsection{Real-world experiments }
\label{subsec:Real_word_exp}

\subsubsection{Real-world datasets and structural regimes}
\label{sec:data_choice}

Throughout the remainder of this section, we use the term Base Clustering Method (\BCM) to denote any algorithm that produces a partition of the data according to a fixed objective or construction rule.
The output labels produced by the \BCM{} are used as inputs to \CoHiRF{}. Importantly, \CoHiRF{} does not rely on distances, similarities, or the geometry of the data itself. It does not perform consensus on distances or affinity matrices, but operates exclusively on the labels produced by the \BCM{}, without modifying its objective, optimization procedure, or internal representations

We consider a collection of real-world datasets selected to cover a range of data regimes commonly encountered in clustering applications. Rather than aiming for an exhaustive benchmark, our objective is to analyze how the behavior of \CoHiRF{} depends on the interaction between data characteristics and the choice of the base clustering method (\BCM{}).

The selected datasets span different combinations of sample size, dimensionality, feature type and class structure. They are used to study how these characteristics influence the structure of the resulting clustering outputs under different base clustering methods, within a hierarchical consensus framework. Table~\ref{tab:real_datasets} summarizes their main characteristics.

\begin{table}[ht]
\centering
\caption{Real-world datasets considered in our experiments.}
\label{tab:real_datasets}
\begin{tabular}{lrrrrrc}
\toprule
Dataset & OpenML ID & $n$ & $p$ & $C$ & $p_{cat}$ \\
\midrule
\texttt{alizadeh-2000-v2} & 46773 & 62 & 2094 & 3 & 1 \\
\texttt{garber-2001} & 46779 & 66 & 4554 & 4 & 1 \\
\texttt{shuttle} & 40685 & 58000 & 10 & 7 & 1 \\
\texttt{coil-20} & 46783 & 1440 & 1025 & 0 & 1 \\
\texttt{nursery} & 1568 & 12958 & 9 & 4 & 9 \\
\bottomrule
\end{tabular}
\end{table}

\subsubsection{Base clustering methods (\BCM)}
\label{subsec:BCM_choice}

The \BCM{} considered in our experiments belong to different methodological families and rely on distinct principles to produce cluster assignments. Rather than assuming a priori which methods are best suited to a given dataset, we use this diversity to analyze how the interaction between data characteristics and base clustering behavior affects the outcome of hierarchical consensus.

Below, we briefly summarize the main properties of the base methods used in our experiments.

\textbf{(i) K-Means} \citep{macqueen1967} is a distance-based clustering method that partitions the data by minimizing the within-cluster sum of squared Euclidean distances to cluster centroids. Its formulation relies on a global objective defined over all samples and clusters and is naturally suited to convex, approximately spherical clusters.
\textbf{(ii) Kernel K-Means} \citep{dhillon2004} extends K-Means by implicitly mapping the data into a higher-dimensional feature space through a kernel function, here approximated using random Fourier features. This allows the method to capture non-linear cluster structures while retaining a global optimization objective.
\textbf{(iii) DBSCAN} \citep{ester1996density} is a density-based clustering method that defines clusters as connected components of regions with sufficient local point density, based on neighborhood relations in the input space. It does not require the number of clusters to be specified in advance and can recover clusters of arbitrary shape when a meaningful notion of neighborhood is available.
\textbf{(iv) SC-SRGF} \citep{liu2020} is a graph-based spectral clustering method that constructs a sparse random graph and applies spectral filtering to approximate spectral embeddings without explicit eigen-decomposition, improving scalability compared to classical spectral clustering.

\textbf{Specificity of SC-SRGF and experimental protocol} SC-SRGF exhibits two properties that directly affect how hierarchical consensus is applied.
First, it is intrinsically stochastic: the sparse random graph construction and graph filtering procedure introduce variability across independent runs, even on the same dataset. Second, SC-SRGF is designed to operate directly in the  feature space and does not rely on random feature subsampling to remain tractable. Accordingly, when SC-SRGF is used as the \BCM{}, we fix $q=p$ and do not apply feature subsampling.

In this setting, \CoHiRF{} does not aim to mitigate high-dimensional noise through multiple feature views. Instead, it provides a hierarchical reduction mechanism through medoid selection, optionally combined with consensus across independent realizations of the graph construction.
We consider two fixed values of $R$, which are part of the experimental protocol and are not tuned:\textbf{(i)} $R=1$, where SC-SRGF is applied once per iteration and \CoHiRF{} acts primarily as a hierarchical compression mechanism and \textbf{(ii)} $R=2$, where SC-SRGF is applied twice independently at each iteration and strict consensus retains only label relations that are reproducible across both graph realizations.
We do not consider a relaxed consensus variant in this setting, as variability arises from intrinsic graph randomness rather than from isolated unstable feature views.

More generally, while \CoHiRF{} incurs additional computational cost compared to a single run of the base clustering method, this overhead remains moderate in all experiments considered. Importantly, \CoHiRF{} produces an explicit Cluster Fusion Hierarchy (\CFH), providing an interpretable, multi-resolution representation of cluster relationships that is not available with standard flat clustering methods.

\subsubsection{Results on high-dimensional gene expression datasets}
\label{subsec:gene}

Gene expression microarray datasets such as \texttt{alizadeh-2000-v2} and \texttt{garber-2001} are characterized by a very large number of dimensions and a small number of samples ($p \gg n$). In such settings, distances in the original feature space are known to be affected by the presence of many weakly informative dimensions. However, the original studies \citet{aliz2000} and \citet{garber2001} show that gene expression signatures are associated with distinct biological classes and that meaningful groupings can be recovered in relevant low-dimensional subspaces, as illustrated by the unsupervised analyses reported in these works.

For both datasets, base clustering methods relying on global objectives (K-Means and SC-SRGF) achieve positive ARI values, with substantially higher scores on \texttt{alizadeh-2000-v2} than on \texttt{garber-2001}. For these base methods, the application of \CoHiRF{} is associated with higher ARI values.

For SC-SRGF on \texttt{alizadeh-2000-v2}, when $R=1$, no consensus across repetitions is performed. In this setting, the application of \CoHiRF{} reduces to a hierarchical contraction based on medoid selection and the observed improvement in ARI therefore occurs in the absence of any consensus mechanism.
On \texttt{garber-2001}, performance increases further when moving from $R=1$ to $R=2$, indicating that agreement across independent graph realizations is associated with additional gains.

\begin{table}[ht]
\centering
\caption{Gene expression datasets ($p \gg n$).}
\label{tab:gene_expr}
\begin{subtable}[t]{0.48\linewidth}
\centering
\begin{footnotesize}
\caption{\texttt{alizadeh-2000-v2} $(n,p,c)=(62, 2\,094, 3)$.}
\label{tab:K-Means_vs_cohirf}
\begin{tabular}{llll}
\toprule
 Model & \ARI &   Time (s) \\
\midrule
\textcolor{blue}{KMeans} & 0.838 $\pm$ 0.014 &   \bfseries 0.028 $\pm$ 0.004 \\
\CoHiRF & \underline{0.871 $\pm$ 0.014} &  \underline{0.078 $\pm$ 0.025} \\
\RCoHiRF & \bfseries 0.905 $\pm$ 0.039 &    0.384 $\pm$ 0.915 \\
\midrule
\textcolor{blue}{\KK}& 0.044 $\pm$ 0.015 &  \bfseries 0.096 $\pm$ 0.020 \\
\CoHiRF & \underline{0.083 $\pm$ 0.034} & 0.388 $\pm$ 0.318 \\
\RCoHiRF & \bfseries 0.142 $\pm$ 0.060 &  \underline{0.241 $\pm$ 0.077} \\
\midrule
\textcolor{blue}{DBSCAN}&\bfseries \underline{0.000 $\pm$ 0.000} &  \bfseries 0.072 $\pm$ 0.008 \\
\CoHiRF & \bfseries \underline{0.000 $\pm$ 0.000}  &   \underline{1.891 $\pm$ 0.121} \\
\RCoHiRF  & \bfseries \underline{0.000 $\pm$ 0.000} &  3.367 $\pm$ 0.243 \\
\midrule
\textcolor{blue}{SC-SRGF}& \underline{0.891 $\pm$ 0.125}  & \bfseries 0.539 $\pm$ 0.277 \\
\CoHiRF(1R) & \bfseries 0.947 $\pm$ 0.000 &   2.913 $\pm$ 1.498 \\
\CoHiRF(2R) & \bfseries 0.947 $\pm$ 0.000 & \underline{1.259 $\pm$ 1.089} \\
\bottomrule
\end{tabular}
\end{footnotesize}
\end{subtable}
\hfill
\begin{subtable}[t]{0.48\linewidth}
\centering
\begin{footnotesize}
\caption{\texttt{garber-2001} $(n,p,c)=(66, 4\,554, 4)$}
\label{tab:garber}
\begin{tabular}{llll}
\toprule
 Model & \ARI & Time (s) \\
\midrule
\textcolor{blue}{KMeans}& 0.248 $\pm$ 0.044 &  \bfseries 0.056 $\pm$ 0.011 \\
\CoHiRF & \bfseries 0.292 $\pm$ 0.034 &    \underline{0.245 $\pm$ 0.118} \\
\RCoHiRF & \underline{0.260 $\pm$ 0.053} &  0.336 $\pm$ 0.130 \\
\midrule
\textcolor{blue}{\KK}& \underline{0.054 $\pm$ 0.021} &   \bfseries 0.180 $\pm$ 0.058 \\
\CoHiRF & \underline{0.054 $\pm$ 0.015} &   \underline{0.555 $\pm$ 0.225} \\
\RCoHiRF & \bfseries 0.147 $\pm$ 0.056 &  0.639 $\pm$ 0.806 \\
\midrule
\textcolor{blue}{DBSCAN} & \bfseries \underline{0.000 $\pm$ 0.000} & \bfseries 0.100 $\pm$ 0.007 \\
\CoHiRF & \bfseries \underline{0.000 $\pm$ 0.000} &  \underline{1.974 $\pm$ 0.138} \\
\RCoHiRF  & \bfseries \underline{0.000 $\pm$ 0.000} &  3.419 $\pm$ 0.241 \\
\midrule
\textcolor{blue}{SC-SRGF}& 0.204 $\pm$ 0.035 &  \bfseries 0.553 $\pm$ 0.291 \\
\CoHiRF(1R) & \underline{0.235 $\pm$ 0.013} &  \underline{1.066 $\pm$ 0.814} \\
\CoHiRF(2R) & \bfseries 0.454 $\pm$ 0.034 &  1.700 $\pm$ 1.500 \\
\bottomrule
\end{tabular}
\end{footnotesize}
\end{subtable}
\end{table}

Methods that are poorly aligned with this data regime (e.g., DBSCAN and Kernel K-Means) yield ARI values close to zero on both datasets. In these cases, applying \CoHiRF{} does not lead to meaningful improvements.

\subsubsection{Results on large-sample, low-dimensional data (Statlog Shuttle)}
\label{subsec:shuttle}

The \texttt{shuttle} dataset (Statlog Shuttle) corresponds to a regime with a large number of samples and a small number of numerical features ($p \ll n$), as documented in the UCI Machine Learning Repository \citep{DuaGraff2019}. It contains seven highly imbalanced classes, with one dominant class accounting for the majority of observations. The dataset has been widely used as a benchmark in the StatLog project and in subsequent machine learning studies, indicating that the class labels are meaningfully related to the original feature space \citep{KingEtAl1995}.

\begin{table}[ht]
\centering
\begin{footnotesize}
\caption{\texttt{shuttle} $(n,p,c)=(58\,000, 10, 7)$.}
\label{tab:shuttle}
\begin{tabular}{llll}
\toprule
 Model & \ARI &  Time (s) \\
\midrule
  \textcolor{blue}{KMeans} & \underline{0.608 $\pm$ 0.000} &    \bfseries 0.093 $\pm$ 0.053 \\
  CoHiRF & 0.579 $\pm$ 0.018 &    15.860 $\pm$ 5.240 \\
 \RCoHiRF & \bfseries 0.646 $\pm$ 0.025 &    20.328 $\pm$ 3.065 \\
\midrule

  \textcolor{blue}{\KK}& 0.250 $\pm$ 0.084 &    \bfseries 1.701 $\pm$ 0.241 \\
 \CoHiRF & \bfseries 0.581 $\pm$ 0.114 &   19.815 $\pm$ 12.709 \\
\RCoHiRF & \underline{0.553 $\pm$ 0.074} &   32.912 $\pm$ 17.803 \\
\midrule

 \textcolor{blue}{DBSCAN} &\bfseries  0.709 $\pm$ 0.013 &  \bfseries  36.204 $\pm$ 18.613 \\
\CoHiRF & \underline{0.704 $\pm$ 0.016} & \underline{ 62.587 $\pm$ 31.710} \\
  \RCoHiRF & 0.696 $\pm$ 0.015 &   63.722 $\pm$ 28.394 \\
\midrule

 \textcolor{blue}{SC-SRGF} & \multicolumn{2}{c}{\textit{infeasible (under our memory limit)}} \\
 \CoHiRFb(1R) & \bfseries \underline{0.456 $\pm$ 0.017} &  \bfseries \underline{286.274 $\pm$ 63.310} \\

\bottomrule
\end{tabular}
\end{footnotesize}
\end{table}

In this setting, K-Means already achieves strong performance according to both external and internal validation metrics. Applying \CoHiRF{} to K-Means yields comparable results overall, with only limited variation between strict and relaxed consensus variants. A similar behavior is observed for DBSCAN, whose performance remains essentially unchanged when combined with \CoHiRF{}, indicating that hierarchical contraction does not substantially modify the resulting partition in this regime.

For Kernel K-Means, the application of \CoHiRF{} produces substantially different partitions, reflected by a marked increase in ARI compared to the base method. This shows that hierarchical aggregation and consensus can significantly affect the output of kernel-based methods on this dataset, without implying that such changes systematically improve or degrade clustering quality across base methods.

Finally, SC-SRGF cannot be executed directly on this dataset under our memory constraints. Its batched variant, \CoHiRFb{}, enables spectral clustering to be applied at this scale, yielding a valid partition where the base method alone is infeasible. This result highlights the role of \CoHiRF{} as an enabling mechanism for large-scale clustering, independently of its effect on clustering accuracy.

\subsubsection{Results on moderate-sample, high-dimensional image data (COIL-20)}
\label{subsec:coil}


The \texttt{coil-20} dataset consists of grayscale images of 20 objects acquired under systematically varying viewing angles ($p \approx n$), as introduced in the Columbia Object Image Library \citep{NeneEtAl1996}. For each object, images are collected at regular angular increments over a full rotation, resulting in a fixed number of views per object. The dataset has been widely used in studies on dimensionality reduction and representation learning, in which images are represented as high-dimensional vectors derived directly from pixel intensities \citep{TenenbaumEtAl2000}. The \texttt{coil-20} dataset is particularly useful for studying the behavior of clustering algorithms, as the structure of the resulting outputs can vary substantially depending on the underlying base clustering method (BCM).

\begin{table}[ht]
\centering
\begin{footnotesize}
\caption{\texttt{coil-20} $(n,p,c)=(1\,440, 1\,025, 20)$.}
\label{tab:coil}
\begin{tabular}{llll}
\toprule
Model & \ARI &  Time (s) \\
\midrule
\textcolor{blue}{KMeans} & \bfseries 0.664 $\pm$ 0.005 & \bfseries 0.099 $\pm$ 0.050 \\
\CoHiRF & \underline{0.335 $\pm$ 0.015} &  \underline{0.327 $\pm$ 0.119} \\
\RCoHiRF & 0.244 $\pm$ 0.021 &  0.726 $\pm$ 0.878 \\
\midrule

\textcolor{blue}{\KK}& 0.004 $\pm$ 0.003 &  \bfseries 0.131 $\pm$ 0.046 \\
\CoHiRF & \bfseries 0.018 $\pm$ 0.028 &0.871 $\pm$ 1.071 \\
\RCoHiRF & \underline{0.012 $\pm$ 0.015} & \underline{0.472 $\pm$ 0.113} \\
\midrule

\textcolor{blue}{DBSCAN} & 0.143 $\pm$ 0.000 &  \bfseries 0.088 $\pm$ 0.009 \\
\CoHiRF & \bfseries 0.725 $\pm$ 0.021 &   \underline{1.215 $\pm$ 0.563} \\
\RCoHiRF  & \underline{0.721 $\pm$ 0.024} & 1.379 $\pm$ 0.485 \\
\midrule

\textcolor{blue}{SC-SRGF}& \bfseries 0.775 $\pm$ 0.005 &  89.340 $\pm$ 46.321 \\
\CoHiRF(1R) & 0.164 $\pm$ 0.013 & \bfseries 57.324 $\pm$ 75.007 \\
\CoHiRF(2R) & \underline{0.200 $\pm$ 0.014} & \underline{65.973 $\pm$ 47.384} \\

\bottomrule
\end{tabular}
\end{footnotesize}
\end{table}

These results indicate that hierarchical contraction interacts differently with graph-based and density-based clustering methods.

SC-SRGF relies on a graph representation of the data, where cluster assignments depend on connectivity patterns induced by neighborhood relations. On COIL-20, images of the same object correspond to smoothly varying viewpoints, which induce locally continuous neighborhood relations in the feature space. SC-SRGF exploits these relations through its graph-based formulation. Hierarchical contraction modifies the set of nodes used to build the graph by replacing groups of samples with representatives, which can modify local connectivity patterns on which SC-SRGF depends. This sensitivity is reflected in the observed degradation of performance when SC-SRGF is combined with hierarchical contraction.

DBSCAN, by contrast, does not aim to recover chain-like or globally continuous relational structures, but identifies clusters as connected regions of sufficient local density. It relies on local density connectivity and does not explicitly enforce the preservation of global connectivity relations. On COIL-20, this behavior is not well aligned with the underlying data structure, which is characterized by smoothly varying viewpoints inducing continuous neighborhood relations and DBSCAN alone does not recover meaningful partitions. When combined with \CoHiRF{}, hierarchical contraction aggregates locally consistent density-based groups through medoid selection across iterations. Empirically, this process yields partitions that are more consistent with the object-level structure of the dataset, as reflected by higher ARI values.

K-Means partitions the data by minimizing within-cluster variance in the feature space and does not exploit neighborhood or continuity relations.
On COIL-20, images of the same object lie on a smooth, low-dimensional manifold induced by gradual viewpoint changes. Random feature projections disrupt this structure by mixing pixel intensities in a way that does not preserve local continuity along the manifold. As a result, K-Means applied to projected feature spaces yields unstable or misaligned partitions. Applying hierarchical contraction further replaces sets of samples by representative medoids, reducing the sampling density along the manifold. In this setting, contraction does not preserve the geometric structure relevant to K-Means objective and is associated with a further degradation of clustering performance.

Kernel K-Means operates by implicitly mapping the data into a high-dimensional feature space defined by the chosen kernel, here approximated using random Fourier features. The resulting similarity structure depends on global pairwise relations induced by the kernel, rather than on explicit local neighborhood continuity. On COIL-20, images of the same object form smooth trajectories corresponding to gradual viewpoint changes. Kernel-induced feature mappings do not explicitly preserve this sequential or manifold-like structure and instead emphasize global similarities that may not align with the underlying viewpoint continuity. As a result, Kernel K-Means yields partitions with very low ARI on this dataset. Applying \CoHiRF{} to Kernel K-Means operates exclusively on these label assignments. Since the base partitions do not reflect the object-level structure, hierarchical contraction and consensus do not recover meaningful clusters and only marginally affect performance.

\subsection{Empirical insights on hierarchical consensus and contraction}
\label{sec:Empirical_insights}

Across the synthetic and real-world experiments, a consistent empirical pattern emerges regarding when \CoHiRF{} is beneficial.

\CoHiRF{} tends to be effective when the base clustering method produces label relations that are\\
\textbf{(i)} reproducible across multiple runs of the base method applied to different feature views or stochastic realizations and\\
\textbf{(ii)} compatible with representative-based contraction, in the sense that replacing groups of samples by medoids preserves the relational structure that is relevant for the base method.

In practice, reproducibility is observed either under random feature projections, as in centroid-based methods, or under independent stochastic graph constructions,
as in graph-based methods such as SC-SRGF. When these conditions are satisfied, hierarchical contraction and consensus reinforce stable label relations, leading to improved robustness, increased stability, or extended scalability. When they are not, contraction may have a neutral effect or degrade performance by altering structural properties on which the base method relies. We emphasize that this observation is purely empirical and does not constitute a guarantee of performance.
It should be understood as a unifying interpretation of the experimental results, rather than as a prescriptive rule for method selection.

A useful way to interpret this empirical behavior is through an analogy with \textbf{margin-based robustness}, commonly used in supervised learning.
In that context, points far from decision boundaries (high-margin) exhibit stable predictions under repeated perturbations of the training process, while points near boundaries (low-margin) are sensitive to small variations. A similar distinction appears to arise in clustering. Some label relations are consistently reproduced across feature projections or stochastic realizations of the base method, while others vary across runs. By enforcing agreement across multiple runs, \CoHiRF{} preserves label relations that are consistently reproduced and suppresses unstable ones. Hierarchical aggregation then propagates these stable relations across scales.

We stress that this margin-based perspective is conceptual rather than formal: \CoHiRF{} does not define, estimate, or optimize any explicit notion of margin. Nevertheless, it provides an intuitive lens through which the interaction between consensus, contraction and base clustering behavior can be understood and suggests promising directions for future theoretical analysis.

\section{Conclusion}
\label{sec:conclusion}

In this work, we introduced \CoHiRF{}, a hierarchical consensus framework for clustering that operates exclusively at the level of label assignments produced by an arbitrary base clustering method. Rather than proposing a new clustering objective, \CoHiRF{} acts as a meta-algorithm: it repeatedly applies a base method to multiple views or realizations, identifies and retains consistent label relations through consensus and progressively reduces the problem size via representative-based contraction. This design makes it possible to extend the applicability of existing clustering algorithms beyond their usual computational limits, while preserving their underlying modeling assumptions.

Across a broad set of synthetic and real-world experiments, we showed that \CoHiRF{} can substantially improve robustness to high-dimensional noise, enhance stability under stochastic variability and enable scalability to regimes where the base method alone is infeasible. In particular, we demonstrated how hierarchical contraction allows graph-based and density-based methods to scale to very large sample sizes and how consensus across multiple views mitigates instability induced by non-informative features. Beyond the final flat partition, \CoHiRF{} produces an explicit Cluster Fusion Hierarchy, providing a multi-resolution and interpretable representation of the clustering process.

Our experimental study also highlights that hierarchical consensus and contraction do not universally improve clustering performance. Instead, their effectiveness depends on the interaction between the base clustering method and the structure of the data. When the base method produces label relations that are reproducible across runs and compatible with representative-based contraction, \CoHiRF{} reinforces these stable relations and yields empirical gains. When these conditions are not met, contraction may have a neutral effect or alter structural properties on which the base method relies. We emphasize that this observation is empirical in nature and should be understood as a unifying interpretation of our results rather than as a prescriptive rule.

Several directions for future work emerge from this study. From a theoretical perspective, formalizing the notion of stability underlying hierarchical consensus, potentially through margin-based or robustness analyses, remains an open challenge. From a practical standpoint, adaptive strategies for selecting consensus strength, contraction schedules, or batching policies could further improve performance across diverse data regimes. More broadly, we believe that hierarchical consensus provides a flexible and interpretable framework for revisiting large-scale clustering beyond the dominance of centroid-based methods.



\bibliography{CohirfBIB}

@inproceedings{rahimi2007,
  title={Random Features for Large-Scale Kernel Machines},
  author={Rahimi, Ali and Recht, Benjamin},
  booktitle={Advances in Neural Information Processing Systems},
  year={2007}
}

@article{NeneEtAl1996,
  author  = {Nene, Sameer A. and Nayar, Shree K. and Murase, Hiroshi},
  title   = {Columbia Object Image Library (COIL-20)},
  journal = {Technical Report CUCS-005-96},
  year    = {1996},
  institution = {Columbia University}
}

@article{TenenbaumEtAl2000,
  author  = {Tenenbaum, Joshua B. and de Silva, Vin and Langford, John C.},
  title   = {A Global Geometric Framework for Nonlinear Dimensionality Reduction},
  journal = {Science},
  volume  = {290},
  number  = {5500},
  pages   = {2319--2323},
  year    = {2000}
}

@misc{DuaGraff2019,
  author       = {Dua, Dheeru and Graff, Casey},
  title        = {UCI Machine Learning Repository},
  year         = {2019},
  howpublished = {\url{https://archive.ics.uci.edu/ml}},
  note         = {Dataset: Statlog (Shuttle)},
  institution  = {University of California, Irvine}
}

@article{KingEtAl1995,
  author  = {King, Ross D. and Feng, Chao and Sutherland, Alistair},
  title   = {Statlog: Comparison of classification algorithms on large real-world problems},
  journal = {Applied Artificial Intelligence},
  volume  = {9},
  number  = {3},
  pages   = {289--333},
  year    = {1995}
}

@article{garber2001,
author = {Garber, Mitchell and Troyanskaya, Olga and Schluens, Karsten and Petersen, Simone and Thaesler, Zsuzsanna and Pacyna-Gengelbach, Manuela and Rijn, Matt and Rosen, Glenn and Perou, Charles and Whyte, Richard and Altman, Russ and Brown, Patrick and Botstein, David and Petersen, Iver},
year = {2001},
month = {12},
pages = {13784-9},
title = {Diversity of gene expression in adenocarcinoma of the lung},
volume = {98},
journal = {Proceedings of the National Academy of Sciences of the United States of America},
doi = {10.1073/pnas.241500798}
}

@article{aliz2000,
	author = {Alizadeh, Ash A. and Eisen, Michael B. and Davis, R. Eric and Ma, Chi and Lossos, Izidore S. and Rosenwald, Andreas and Boldrick, Jennifer C. and Sabet, Hajeer and Tran, Truc and Yu, Xin and Powell, John I. and Yang, Liming and Marti, Gerald E. and Moore, Troy and Hudson, James and Lu, Lisheng and Lewis, David B. and Tibshirani, Robert and Sherlock, Gavin and Chan, Wing C. and Greiner, Timothy C. and Weisenburger, Dennis D. and Armitage, James O. and Warnke, Roger and Levy, Ronald and Wilson, Wyndham and Grever, Michael R. and Byrd, John C. and Botstein, David and Brown, Patrick O. and Staudt, Louis M.},
	date = {2000/02/01},
	doi = {10.1038/35000501},
	id = {Alizadeh2000},
	isbn = {1476-4687},
	journal = {Nature},
	number = {6769},
	pages = {503--511},
	title = {Distinct types of diffuse large B-cell lymphoma identified by gene expression profiling},
	url = {https://doi.org/10.1038/35000501},
	volume = {403},
	year = {2000}}

@article{calinsky1974,
  title={A dendrite method for cluster analysis},
  author={Calinski, Tadeusz and Harabasz, Jerzy},
  journal={Communications in Statistics},
  volume={3},
  number={1},
  pages={1--27},
  year={1974}
}

@inproceedings{macqueen1967,
  author = {MacQueen, J.},
  title = {Some methods for classification and analysis of multivariate observations},
  booktitle = {Proceedings of the Fifth Berkeley Symposium on Mathematical Statistics and Probability},
  volume = {1},
  pages = {281--297},
  year = {1967}
}

@inproceedings{ng2001,
  author = {Ng, Andrew Y. and Jordan, Michael I. and Weiss, Yair},
  title = {On spectral clustering: Analysis and an algorithm},
  booktitle = {Advances in Neural Information Processing Systems (NIPS)},
  pages = {849--856},
  year = {2001}
}

@article{vonluxburg2007,
  author = {von Luxburg, Ulrike},
  title = {A tutorial on spectral clustering},
  journal = {Statistics and Computing},
  volume = {17},
  number = {4},
  pages = {395--416},
  year = {2007}
}

@article{scholkopf1998,
  author = {Sch{\"o}lkopf, Bernhard and Smola, Alexander and M{\"u}ller, Klaus-Robert},
  title = {Nonlinear component analysis as a kernel eigenvalue problem},
  journal = {Neural Computation},
  volume = {10},
  number = {5},
  pages = {1299--1319},
  year = {1998}
}

@article{dhillon2004,
  author = {Dhillon, Inderjit S. and Guan, Yuqiang and Kulis, Brian},
  title = {Kernel k-means: spectral clustering and normalized cuts},
  journal = {Proceedings of the Tenth ACM SIGKDD International Conference on Knowledge Discovery and Data Mining},
  pages = {551--556},
  year = {2004}
}

@article{kiselev2019,
  author = {Kiselev, Vladimir Yu and Andrews, Tallulah S. and Hemberg, Martin},
  title = {Challenges in unsupervised clustering of single-cell RNA-seq data},
  journal = {Nature Reviews Genetics},
  volume = {20},
  number = {5},
  pages = {273--282},
  year = {2019}
}

@inproceedings{campello2013,
  author = {Campello, Ricardo J. G. B. and Moulavi, Davoud and Sander, J{\"o}rg},
  title = {Density-based clustering based on hierarchical density estimates},
  booktitle = {Pacific-Asia Conference on Knowledge Discovery and Data Mining (PAKDD)},
  pages = {160--172},
  year = {2013}
}

@inproceedings{bradley1998,
  author = {Bradley, Paul S. and Fayyad, Usama M. and Reina, Cory},
  title = {Scaling clustering algorithms to large databases},
  booktitle = {Proceedings of the Fourth International Conference on Knowledge Discovery and Data Mining (KDD)},
  pages = {9--15},
  year = {1998}
}

@article{farnstrom2000,
  author = {Farnstrom, Fredrik and Lewis, James and Elkan, Charles},
  title = {Scalability for clustering algorithms revisited},
  journal = {ACM SIGKDD Explorations Newsletter},
  volume = {2},
  number = {1},
  pages = {51--57},
  year = {2000}
}

@inproceedings{boutsidis2010,
  author = {Boutsidis, Christos and Mahoney, Michael W. and Drineas, Petros},
  title = {Unsupervised feature selection for the k-means clustering problem},
  booktitle = {Advances in Neural Information Processing Systems (NIPS)},
  pages = {153--161},
  year = {2010}
}

@inproceedings{kumar2004,
  author = {Kumar, Amit and Sabharwal, Yogish and Sen, Sandeep},
  title = {A simple linear time (1 + $\epsilon$)-approximation algorithm for k-means clustering in any dimensions},
  booktitle = {Proceedings of the 45th Annual IEEE Symposium on Foundations of Computer Science (FOCS)},
  pages = {454--462},
  year = {2004}
}

@inproceedings{bachem2016,
  author = {Bachem, Olivier and Lucic, Mario and Krause, Andreas},
  title = {Scalable k-means clustering via lightweight coresets},
  booktitle = {Proceedings of the 22nd ACM SIGKDD International Conference on Knowledge Discovery and Data Mining},
  pages = {1119--1127},
  year = {2016}
}

@inproceedings{zelnik2004,
  author = {Zelnik-Manor, Lihi and Perona, Pietro},
  title = {Self-tuning spectral clustering},
  booktitle = {Advances in Neural Information Processing Systems (NIPS)},
  pages = {1601--1608},
  year = {2004}
}

@article{chen2011,
  author  = {Chen, Wei-Yen and Song, Yangqiu and Bai, Hongjie and Lin, Chih-Jen and Chang, Edward Y.},
  title   = {Parallel spectral clustering in distributed systems},
  journal = {IEEE Transactions on Pattern Analysis and Machine Intelligence},
  volume  = {33},
  number  = {3},
  pages   = {568--586},
  year    = {2011}
}

@article{wang2020,
  author = {Wang, Jianbo and Wang, Xiao and Tian, Fei and Liu, Chang Hong and Yu, Hongkai and Liu, Yue},
  title = {Constrained-size spectral clustering},
  journal = {Knowledge-Based Systems},
  volume = {199},
  pages = {105924},
  year = {2020}
}

@article{liu2020,
  author = {Liu, Bo and Wang, Yanshan and Zhang, Yu-Jin and Smola, Alex},
  title = {Simple and scalable sparse k-means clustering via feature ranking},
  journal = {Proceedings of the AAAI Conference on Artificial Intelligence},
  volume = {34},
  pages = {4946--4953},
  year = {2020}
}

@article{murtagh2012,
  author = {Murtagh, Fionn and Contreras, Pedro},
  title = {Algorithms for hierarchical clustering: an overview},
  journal = {Wiley Interdisciplinary Reviews: Data Mining and Knowledge Discovery},
  volume = {2},
  number = {1},
  pages = {86--97},
  year = {2012}
}

@inproceedings{sculley2010,
  author = {Sculley, D.},
  title = {Web-scale k-means clustering},
  booktitle = {Proceedings of the 19th International Conference on World Wide Web},
  pages = {1177--1178},
  year = {2010}
}

@inproceedings{elkan2003,
  author    = {Elkan, Charles},
  title     = {Using the triangle inequality to accelerate k-means},
  booktitle = {Proceedings of the 20th International Conference on Machine Learning (ICML)},
  pages     = {147--153},
  year      = {2003}
}

@article{monti2003,
  author = {Monti, Stefano and Tamayo, Pablo and Mesirov, Jill and Golub, Todd},
  title = {Consensus clustering: a resampling-based method for class discovery and visualization of gene expression microarray data},
  journal = {Machine Learning},
  volume = {52},
  number = {1},
  pages = {91--118},
  year = {2003}
}

@article{vega2007,
  author = {Vega-Pons, Sandro and Ruiz-Shulcloper, Jos{\'e}},
  title = {A survey of clustering ensemble algorithms},
  journal = {International Journal of Pattern Recognition and Artificial Intelligence},
  volume = {25},
  number = {03},
  pages = {337--372},
  year = {2011}
}

@article{pedregosa2011,
  author = {Pedregosa, Fabian and Varoquaux, Ga{\"e}l and Gramfort, Alexandre and Michel, Vincent and Thirion, Bertrand and Grisel, Olivier and Blondel, Mathieu and Prettenhofer, Peter and Weiss, Ron and Dubourg, Vincent and others},
  title = {Scikit-learn: Machine learning in Python},
  journal = {Journal of Machine Learning Research},
  volume = {12},
  pages = {2825--2830},
  year = {2011}
}

@inproceedings{fern2003random,
  title={Random projection for high dimensional data clustering: A cluster ensemble approach},
  author={Fern, Xiaoli Z and Brodley, Carla E},
  booktitle={Proceedings of the 20th International Conference on Machine Learning (ICML)},
  pages={186--193},
  year={2003}
}

@article{fred2005combining,
  title   = {Combining multiple clusterings using evidence accumulation},
  author  = {Fred, Ana L. N. and Jain, Anil K.},
  journal = {IEEE Transactions on Pattern Analysis and Machine Intelligence},
  volume  = {27},
  number  = {6},
  pages   = {835--850},
  year    = {2005}
}

@article{strehl2002,
  title={Cluster ensembles - a knowledge reuse framework for combining multiple partitions},
  author={Strehl, Alexander and Ghosh, Joydeep},
  journal={Journal of machine learning research},
  volume={3},
  pages={583-617},
  year={2002}
}

@inproceedings{ankerst1999optics,
  title={OPTICS: Ordering points to identify the clustering structure},
  author={Ankerst, Mihael and Breunig, Markus M and Kriegel, Hans-Peter and Sander, J{\"o}rg},
  booktitle={Proceedings of the 1999 ACM SIGMOD international conference on Management of data},
  pages={49-60},
  year={1999},
  organization={ACM}
}

@INPROCEEDINGS{ester1996density,
    author = {{Ester}, Martin and {Kriegel}, Hans-Peter and {Sander}, J{\"o}rg and {Xu}, Xiaowei},
        title = "{A Density-Based Algorithm for Discovering Clusters in Large Spatial Databases with Noise}",
    booktitle = {Second International Conference on Knowledge Discovery and Data Mining (KDD'96). Proceedings of a conference held August 2-4},
         year = 1996,
       editor = {{Pfitzner}, D.~W. and {Salmon}, J.~K.},
        month = jan,
        pages = {226-331},
       adsurl = {https://ui.adsabs.harvard.edu/abs/1996kddm.conf..226E}
}

@article{ari1985,
  title = {Comparing Partitions},
  author = {Hubert, Lawrence and Arabie, Phipps},
  year = {1985},
  month = dec,
  journal = {Journal of Classification},
  volume = {2},
  number = {1},
  pages = {193-218},
  issn = {1432-1343},
  doi = {10.1007/BF01908075},
  urldate = {2025-01-26},
  langid = {english}
}

@inproceedings{beyer1999nearest,
  title = {When {{Is}} "{{Nearest Neighbor}}" {{Meaningful}}?},
  booktitle = {Database {{Theory}} -- {{ICDT}}'99},
  author = {Beyer, Kevin and Goldstein, Jonathan and Ramakrishnan, Raghu and Shaft, Uri},
  editor = {Beeri, Catriel and Buneman, Peter},
  year = {1999},
  pages = {217-235},
  publisher = {Springer},
  address = {Berlin, Heidelberg},
  doi = {10.1007/3-540-49257-7_15},
  isbn = {978-3-540-49257-3},
  langid = {english}
}

@article{silhouette1987,
  title = {Silhouettes: {{A}} Graphical Aid to the Interpretation and Validation of Cluster Analysis},
  shorttitle = {Silhouettes},
  author = {Rousseeuw, Peter J.},
  year = {1987},
  month = nov,
  journal = {Journal of Computational and Applied Mathematics},
  volume = {20},
  pages = {53-65},
  issn = {0377-0427},
  doi = {10.1016/0377-0427(87)90125-7},
  urldate = {2025-01-26}
}

@article{spectralclustering2000,
  title = {Normalized Cuts and Image Segmentation},
  author = {Shi, Jianbo and Malik, J.},
  year = {2000},
  month = aug,
  journal = {IEEE Transactions on Pattern Analysis and Machine Intelligence},
  volume = {22},
  number = {8},
  pages = {888-905},
  issn = {1939-3539},
  doi = {10.1109/34.868688},
  urldate = {2025-01-10}
}

@misc{tpe2023,
  title = {Tree-{{Structured Parzen Estimator}}: {{Understanding Its Algorithm Components}} and {{Their Roles}} for {{Better Empirical Performance}}},
  shorttitle = {Tree-{{Structured Parzen Estimator}}},
  author = {Watanabe, Shuhei},
  year = {2023},
  month = may,
  number = {arXiv:2304.11127},
  eprint = {2304.11127},
  primaryclass = {cs},
  publisher = {arXiv},
  doi = {10.48550/arXiv.2304.11127},
  urldate = {2025-01-27},
  archiveprefix = {arXiv}
}
\bibliographystyle{tmlr}

\appendix

\section{Overall complexity proofs}
\label{app:complexity_proofs}

This appendix provides detailed derivations of the complexity results stated in
Theorems~\ref{thm:complexity}-\ref{thm:batched_complexity}.

\subsection{Complexity of the \GetClusters{} step}
\label{compl:GetClusters}

We analyze the computational complexity of the \GetClusters{} function, which performs
consensus-based clustering at iteration $e$ on the set of $n^{(e-1)}$ active medoids.

(i) At iteration $e$, \GetClusters{} applies the base clustering method (\BCM) independently to
$R$ random feature projections of dimension $q$.
Let $\TimeComplexity{\BCM{}, n, q}$ denote the time required by the base method to cluster
$n$ samples in dimension $q$.
The total cost of this step is therefore
$$
O\left(R \cdot \TimeComplexity{\BCM{}, n^{(e-1)}, q}\right).
$$

(ii) For each active medoid, a code of length $R$ is constructed by concatenating its cluster
labels across all projections.
This requires $O(R)$ operations per medoid, for a total cost of
$$
O(R \,n^{(e-1)}).
$$
Grouping medoids by identical codes can be implemented using hash-based data structures.
Under standard assumptions on hashing, this grouping step also requires
$O(R \,n^{(e-1)})$ time on average. In the absence of hashing, codes can be sorted lexicographically. This yields a worst-case complexity of
$$
O\left(R \,n^{(e-1)} \log\left(n^{(e-1)}\right)\right),
$$
followed by a linear pass to group identical codes.

Combining (i) and (ii), the total complexity of the \GetClusters{} step at iteration $e$
is
$$
O\left(R \cdot \TimeComplexity{\BCM{}, n^{(e-1)}, q}\right)
\quad \text{on average},
$$
and
$$
O\left(
R \cdot \TimeComplexity{\BCM{}, n^{(e-1)}, q}
+ R \,n^{(e-1)} \log\left(n^{(e-1)}\right)
\right)
\quad \text{in the worst case}.
$$

This reformulation replaces the quadratic cost of naive consensus constructions
(e.g., pairwise co-occurrence matrices) by a near-linear procedure in the number of samples.

\paragraph{Example: K-Means as base method.}
For K-Means with $k$ clusters and $I$ iterations, we have
$$
\TimeComplexity{\text{K-Means}, n, q} = O(n \,q \,k \,I).
$$
Substituting into the expression above yields
$$
\TimeComplexity{\GetClusters}
= O(R \,n^{(e-1)} \,q \,k \,I)
\quad \text{on average},
$$
and
$$
O\left(
R \,n^{(e-1)} \,q \,k \,I
+ R \,n^{(e-1)} \log\left(n^{(e-1)}\right)
\right)
\quad \text{in the worst case}.
$$

In practice, the base clustering term dominates, and the consensus overhead remains
linear in the number of active medoids.

\subsection{Complexity of the \ChooseMedoids{} step}
\label{compl:ChooseMedoids}

We analyze the computational complexity of the medoid selection step performed by \ChooseMedoids{} at iteration $e$.
Let $\mathcal{I}_k^{(e)} \subseteq \K^{(e-1)}$ denote the set of active medoids assigned to consensus cluster $k$, with size
$n_k^{(e)} = |\mathcal{I}_k^{(e)}|$.

Computing the medoid exactly requires evaluating all pairwise similarities within each cluster.
Using a normalized inner-product criterion, this entails $O\left((n_k^{(e)})^2 \, p\right)$ operations per cluster.
Aggregating over all clusters yields a total complexity of
$$
\TimeComplexity{\ChooseMedoids}
= O\left(\sum_k (n_k^{(e)})^2 \, p\right),
$$
which, in the worst case where a single cluster contains all active medoids, reduces to
$O\left((n^{(e-1)})^2 \, p\right)$.
Such quadratic behavior is impractical for large-scale settings.

To ensure scalability, we employ a subsampling strategy within each consensus cluster.
Specifically, for cluster $k$, we uniformly sample either a fixed fraction $\rho$ of the points or a maximum number $s_{\max}$ of samples, whichever is smaller.
Let
$$
s = \min(\rho \, n_k^{(e)}, s_{\max})
$$
denote the number of sampled points per cluster.
The medoid is then selected among this subset only.

Under this strategy, the per-cluster computational cost becomes $O(s^2 \, p)$, and the total cost over all consensus clusters is
$$
\TimeComplexity{\ChooseMedoids}
= O(n^{(e)} \, s^2 \, p),
$$
where $n^{(e)}$ denotes the number of consensus clusters.
When $s_{\max}$ is fixed, this complexity simplifies to $O(n^{(e)} \, p)$, which is linear in the number of consensus clusters.

Empirically, this approximation introduces only minor differences in the resulting hierarchy while significantly reducing computational and memory requirements.
As a result, the medoid selection step remains tractable even for large datasets.

\subsection{Complexity of the \UpdateParents{} step}
\label{compl:UpdateParents}

The \UpdateParents{} function updates the parent vector $\vP$ by scanning the set of active medoids at iteration $e$.
Parent pointers are assigned according to the consensus clusters identified at that iteration.
The computational cost of this step therefore scales linearly with the number of active medoids,
$$
\TimeComplexity{\UpdateParents} = O(n^{(e-1)}).
$$
Since the number of active medoids decreases across iterations, this cost rapidly becomes negligible compared to the clustering and medoid selection steps.

\subsection{Complexity of the \GetFinalLabels{} step}
\label{compl:GetFinalLabels}

The final labeling step propagates labels through the parent vector $\vP$ to identify the root medoid associated with each sample.
This operation can be implemented using a standard Union-Find (disjoint set union) structure with path compression.
The total computational cost scales linearly with the number of samples,
$$
\TimeComplexity{\GetFinalLabels} = O(n).
$$

and is negligible compared to the iterative clustering and medoid selection steps.

\subsection{Complexity of the relaxed consensus}
\label{compl:RelaxedConsensus}

At iteration $e$, the relaxed consensus operates on at most $R$ repetitions.
For each repetition $r$, a leave-one-out agreement score $\ARI_{\mathrm{LOO}}(r)$ is computed between two partitions of size $n^{(e-1)}$, which requires $O(n^{(e-1)})$ operations.

In the worst case, the procedure removes one repetition at a time, leading to at most $R$ iterations, each involving up to $R$ ARI computations.
The overall complexity of the relaxed consensus step is therefore
$$
\TimeComplexity{\RCoHiRF} = O(R^2 \, n^{(e-1)}).
$$

Since the number of repetitions $R$ is fixed and small in practice, this cost remains linear in the number of active medoids and is negligible compared to the cost of applying the base clustering method.

\subsection{Proof of Theorem~\ref{thm:complexity}}
\label{app:proof:thm:complexity}

We prove the overall computational complexity of \CoHiRF{} by analyzing the cost incurred
at each iteration and summing over all iterations until convergence. At iteration $e$, the algorithm operates on $n^{(e-1)}$ active medoids.
The total computational cost at this iteration is the sum of the costs of the three main components:

(i) Applying the base clustering method to $R$ random projections and constructing the code-based strict consensus requires
$
O\left(R \cdot \TimeComplexity{\BCM{}, n^{(e-1)}, q}\right)
$
time on average (Appendix~\ref{compl:GetClusters}).  This step dominates the cost whenever the base clustering method is at least linear in the number of samples.\\

(ii) As detailed in Appendix~\ref{compl:ChooseMedoids}, medoid selection is performed in the original feature space $\mathbb{R}^p$. With subsampling, its cost is
$
O\left(n^{(e)} \, p\right),
$
where $n^{(e)}$ denotes the number of consensus clusters at iteration $e$.
Since $n^{(e)} \leq n^{(e-1)}$, this cost is asymptotically dominated by the consensus step
whenever the base clustering method is at least linear in the number of samples.\\

(iii) Updating the parent vector requires a single pass over the active medoids and costs
$
O(n^{(e-1)})
$
time (Appendix~\ref{compl:UpdateParents}), which is negligible compared to the
previous terms.

Combining (i)--(ii), the dominant cost at iteration $e$ is therefore
$$
\TimeComplexity{\text{iteration } e}
= O\left(R \cdot \TimeComplexity{\BCM{}, n^{(e-1)}, q}\right).
$$
Summing over the $E$ iterations until convergence yields
$$
\TimeComplexity{\CoHiRF}
= O\!\left(R \cdot \sum_{e=1}^{E} \TimeComplexity{\BCM{}, n^{(e-1)}, q}\right),
$$

Assume that the number of active medoids \textbf{decreases geometrically across iterations}, i.e.,
$
n^{(e)} \leq \alpha \,n^{(e-1)} \quad \text{for some } \alpha < 1.
$
Then $n^{(e)} = \alpha^e n$ and
$$
\sum_{e=1}^{E}n^{(e-1)}= 
\sum_{e=0}^{E-1} n^{(e)}
= n \sum_{e=0}^{E-1} \alpha^e
= O(n).
$$
If the base clustering method has linear complexity in the number of samples, this yields
$$
\TimeComplexity{\CoHiRF}
= O\left(R \cdot \TimeComplexity{\BCM{}, n, q}\right),
$$
which establishes the average-case result stated in Theorem~\ref{thm:complexity}. 
In the \textbf{slow-convergence regime} (worst case), where only a constant number $c>0$ of medoids merge at each iteration,$n^{(e)} = n^{(e-1)} - c$ for some constant $c > 0$,  we have $n^{(e)} = n - e c$ and $E = O(n)$. In this case,
$$
\sum_{e=1}^{E}n^{(e-1)}
= \sum_{e=0}^{E-1} (n - e c)
= O(n^2),
$$
leading to an overall quadratic time complexity.\\

The final labeling step propagates labels through the parent vector and is implemented using a Union-Find structure.
As shown in Appendix~\ref{compl:GetFinalLabels}, this step requires $O(n)$ time and is negligible compared to the iterative clustering cost.
Combining the above results proves Theorem~\ref{thm:complexity}.

\subsection{Proof of Theorem \ref{thm:batched_complexity}}
\label{app:proof:thm:batched_complexity}

We analyze the computational complexity of the batched extension \CoHiRFb{}.
Let $b$ denote the batch size, $B = \lceil n / b \rceil$ the number of batches,
$R$ the number of random projections, and $q$ the projection dimension.
Let $\TimeComplexity{\BCM{}, m, q}$ denote the time required by the base clustering method
to process $m$ samples in dimension $q$. Processing a single batch through one iteration of \CoHiRF{} consists of: (i) applying the base clustering method to $R$ random feature views of size $b \times q$, (ii) constructing the consensus, and (iii) performing medoid selection.
As shown in Appendix~\ref{compl:GetClusters}--\ref{compl:ChooseMedoids}, the dominant cost arises from the base clustering step. The time required to process one batch is therefore
$$
\TimeComplexity{\text{batch}}
= O\left(R \cdot \TimeComplexity{\BCM{}, b, q}\right).
$$
When the $B$ batches are processed \textbf{sequentially}, the total time complexity is
$$
\TimeComplexity{\text{batched, seq}}
= O\left(B \cdot \TimeComplexity{\text{batch}}\right)
= O\left(B \,R \cdot \TimeComplexity{\BCM{}, b, q} \right).
$$
Assuming that the base clustering method has linear complexity in the number of samples,
i.e., $\TimeComplexity{\BCM{}, b, q} = O(b \cdot g(q))$,
the sequential batched complexity simplifies to
$$
\TimeComplexity{\text{batched, seq}} = O(R \cdot n \cdot g(q)),
$$
which matches the non-batched formulation up to a constant factor.

When batches are processed \textbf{in parallel} and \textbf{assuming a geometric reduction}
in the number of active medoids across batching levels,
the number of hierarchical levels satisfies

$$
L = O(\log_b (n)).
$$
The parallel time complexity is then
$$
\TimeComplexity{\text{batched, parallel}}
= O\left(L \cdot \TimeComplexity{\text{batch}}\right)
= O\left(\log_b n \,R \cdot \TimeComplexity{\BCM{}, b, q}\right).
$$

\paragraph{Example: K-Means as base method.}
For K-Means with $k$ clusters and $I$ iterations per projection, we have
$$
\TimeComplexity{\BCM{}, b, q} = O(b \,q \,k \,I).
$$
Substituting into the expressions above yields: 
$
\TimeComplexity{\text{batch}} = O(R \,b \,q \,k \,I),
$ 
a sequential complexity
$
O(R \,n \,q \,k \,I),
$ 
and a parallel  time
$
O(\log_b (n) \,R \,b \,q \,k \,I).
$ 
Although batching introduces additional overhead due to random projections,
consensus construction, and hierarchical aggregation,
it preserves linear scaling in $n$.

\subsection{Proof of Theorem~\ref{thm:space_complexity}}
\label{app:proof:thm:space_complexity}


The proof follows by accounting for the memory required by the data representation,
the projected views, the consensus encoding, and the base clustering method.

\subsubsection{Memory complexity in the standard (non-batched) setting}

In the \textbf{standard (non-batched)} setting , the algorithm operates on the full dataset of $n$ samples.

(i) The input data matrix $X \in \mathbb{R}^{n \times p}$ must be stored in memory, requiring $O(n \,p)$ space.

(ii) At each iteration, $R$ random feature views of the $n^{(e-1)}$ active medoids are constructed. Storing these projected representations requires $O(R \, n^{(e-1)} \, q)$ memory. In addition, the consensus encoding stores, for each active medoid, a code of length $R$ and an associated label, which together require   $O(R \, n^{(e-1)})$ memory. Since $n^{(e-1)} \leq n$ for all iterations, these terms are upper bounded by $O(R \, n \, q)$.

(iii) Let $\text{Memory}(\BCM{}, n^{(e-1)}, q)$ denote the memory required by the base clustering method to process $n^{(e-1)}$ samples in dimension $q$. This memory is allocated during the clustering step at each iteration and represents the dominant method-dependent memory cost.

(iv) The parent vector $\vP \in \llbracket n \rrbracket^{n}$ storing the hierarchical relationships requires $O(n)$ memory. This term is negligible compared to the storage of the data matrix and projected representations.

Combining (i)--(iv), the peak memory usage of the non-batched formulation is
$$
O\left(n \, p+ R \, n \, q  +\text{Memory}(\BCM{}, n, q)\right),
$$
which proves the first part of Theorem~\ref{thm:space_complexity}.

The medoid selection step requires an additional $O(s^2)$ memory to store pairwise similarities within each consensus cluster. This cost corresponds to a temporary similarity matrix of size $s\times s$, where $s$ is a fixed upper bound on the number of sampled points per cluster.
Since $s$ is independent of the dataset size and small in practice ($s=10^3$),
this memory cost is dominated by the storage of the projected data $O(R \, n \, q)$
and the original data matrix $O(n \, p)$ and does not affect the asymptotic space complexity.

\subsubsection{Memory complexity in the batched setting}

We now consider \textbf{the batched extension} \CoHiRFb{} with batch size $b$. In this setting, the algorithm processes one batch at a time and never requires access to all $n$ samples simultaneously.

(i) At any time, only a single batch of size $b$ is loaded in memory, requiring
$O(b \, p)$ space.

(ii) For each batch, $R$ random projections of dimension $q$ are stored, together with the associated consensus codes and labels. This requires
$O(R \, b \, q)$ memory.

(iii) The base clustering method operates on batches of size $b$ and dimension $q$, requiring $\text{Memory}(\BCM{}, b, q)$ memory.

(iv) The parent vector $\vP$ of size $n$ is maintained throughout the algorithm. Its memory cost remains $O(n)$ and does not affect the peak memory usage,
as it can be stored independently of batch processing.

Combining (i)--(iv), the peak memory usage of the batched formulation is
$$
O\left( b \,p +R \,b \,q+\text{Memory}(\BCM{}, b, q)\right),
$$
which is independent of the total dataset size $n$. The above analysis establishes the memory bounds stated in Theorem~\ref{thm:space_complexity} and shows that the batched formulation
of \CoHiRF{} enables clustering under strict memory constraints while
preserving the algorithmic structure.

\section{Practical considerations and hyperparameter guidelines}
\label{app:practical}

\subsection{Hyperparameter selection}
\label{app:algo-search-space}
The hyperparameter search spaces of the Base Clustering Methods (BCM), whether employed as standalone algorithms or within the \CoHiRF{} framework, is detailed in \cref{tab:bcm-hpo-search-space}. These search spaces were defined based on the maximum number of clusters observed in the selected real-world datasets and on practical guidelines provided in the original publications or reference implementations of the respective methods.

\begin{table}[h]
\centering
\caption{Hyperparameter search spaces for Base Clustering Methods (BCM) used in our experiments.}
\label{tab:bcm-hpo-search-space}
\begin{tabular}{@{}lll@{}}
\toprule
Model                            & Parameter                   & Values           \\ \midrule
\multirow{2}{*}{DBSCAN}          & $\epsilon$                  & Float[0.1, 10.0] \\
                                 & Minimum number of samples   & Int[2, 50]       \\ \midrule
KMeans                           & Number of clusters          & Int[2, 30]       \\ \midrule
\multirow{2}{*}{KernelRBFKMeans} & Number of clusters          & Int[2, 30]       \\
                                 & $\gamma$                    & Float[0.1, 30.0] \\ \midrule
\multirow{3}{*}{SC-SRGF}         & Number of affinity matrices & Int[10, 30]      \\
                                 & Sampling ratio              & Float[0.2, 0.8]  \\
                                 & Number of clusters          & Int[2, 30]       \\ \bottomrule
\end{tabular}
\end{table}

\textit{(Number of feature views).}
Across all experiments, we observed that a small number of feature views is sufficient to extract stable structure, provided that such structure is already present in the underlying method. Empirically, values of $R \in [3,5]$ were consistently chosen as optimal, achieving a favorable balance between robustness and computational cost. Very small values ($R \leq 2$) may fail to capture stable relations and instead overemphasize the medoid-contraction component of \CoHiRF{}, whereas larger values typically exhibit diminishing marginal gains.



\subsection{Computational considerations}

\paragraph{(i)} \textit{($R$-fold computational overhead)}  By construction, \CoHiRF{} incurs an $R$-fold overhead with respect to the base clustering method, since the latter is applied independently to $R$ feature views at each iteration.\\
In practice, this overhead is modest relative to the gain in applicability.
Enabling a method such as spectral clustering to operate on datasets with tens or hundreds of thousands of samples is typically more valuable than marginal speedups of methods that already scale well.\\

\paragraph{(ii)} \textit{($R$-fold parallel computation)} Moreover, the $R$ runs are embarrassingly parallel and can be executed independently across cores or compute nodes, resulting in near-linear wall-clock speedups in multi-core environments.

\paragraph{(iii)} \textit{(Batching and scalability)} For large-scale datasets, batching is essential to control memory usage. This is particularly important for base methods whose memory footprint grows superlinearly with the number of samples.
In our experiments, we employed between 10 and 100 batches, depending on memory constraints. Consequently, the batch sizes were between 10-fold and 100-fold smaller than the total number of samples, which provided an effective compromise between computational efficiency and numerical stability. Batching does not modify the underlying logic of \CoHiRF{}, rather, it enables the extraction of representative medoids that can subsequently be aggregated using the same hierarchical consensus mechanism.

\section{Additional experimental results}
\label{app=experiment}

\begin{table}[ht]
\centering
{\fontsize{7}{10}\selectfont
\caption{\texttt{alizadeh-2000-v2} $(n,p,c)=(62, 2\,094, 3)$.}
\label{tab:alizadeh}
\begin{tabular}{llllll}
\toprule
 Model & \ARI & \CH & \Sil&  Time (s) \\
\midrule
\textcolor{blue}{KMeans} & 0.838 $\pm$ 0.014 & \bfseries 15.151 $\pm$ 0.000 & \underline{0.204 $\pm$ 0.036} &  \bfseries 0.028 $\pm$ 0.004 \\
CoHiRF & \underline{0.871 $\pm$ 0.014} & \underline{15.094 $\pm$ 0.126} & 0.186 $\pm$ 0.006 &   \underline{0.078 $\pm$ 0.025} \\
\RCoHiRF & \bfseries 0.905 $\pm$ 0.039 & \bfseries 15.151 $\pm$ 0.000 & \bfseries 0.221 $\pm$ 0.043 &   0.384 $\pm$ 0.915 \\
\midrule
\textcolor{blue}{\KK}& 0.044 $\pm$ 0.015 & 1.376 $\pm$ 0.129 & \underline{0.014 $\pm$ 0.007} &  \bfseries 0.096 $\pm$ 0.020 \\
CoHiRF & \underline{0.083 $\pm$ 0.034} & \underline{1.569 $\pm$ 0.143} & 0.001 $\pm$ 0.012 &  0.388 $\pm$ 0.318 \\
\RCoHiRF & \bfseries 0.142 $\pm$ 0.060 & \bfseries 2.430 $\pm$ 0.537 & \bfseries 0.041 $\pm$ 0.006 &  \underline{0.241 $\pm$ 0.077} \\
\midrule
\textcolor{blue}{DBSCAN}&\bfseries \underline{0.000 $\pm$ 0.000} & \bfseries \underline{0.000 $\pm$ 0.000} & \bfseries \underline{-1.000 $\pm$ 0.000} &   \bfseries 0.072 $\pm$ 0.008 \\
\CoHiRF & \bfseries \underline{0.000 $\pm$ 0.000} & \bfseries \underline{0.000 $\pm$ 0.000} & \bfseries \underline{-1.000 $\pm$ 0.000} &   \underline{1.891 $\pm$ 0.121} \\
\RCoHiRF  & \bfseries \underline{0.000 $\pm$ 0.000} & \bfseries \underline{0.000 $\pm$ 0.000} & \bfseries \underline{-1.000 $\pm$ 0.000} &  3.367 $\pm$ 0.243 \\
\midrule
\textcolor{blue}{SC-SRGF}& \underline{0.891 $\pm$ 0.125} & \bfseries \underline{12.338 $\pm$ 0.000} & \bfseries \underline{0.194 $\pm$ 0.000} & \bfseries 0.539 $\pm$ 0.277 \\
CoHiRF(1R) & \bfseries 0.947 $\pm$ 0.000 & \bfseries \underline{12.338 $\pm$ 0.000} & \bfseries \underline{0.194 $\pm$ 0.000} &  2.913 $\pm$ 1.498 \\
CoHiRF(2R) & \bfseries 0.947 $\pm$ 0.000 & \bfseries \underline{12.338 $\pm$ 0.000} & \bfseries \underline{0.194 $\pm$ 0.000} &   \underline{1.259 $\pm$ 1.089} \\
\bottomrule
\end{tabular}
}
\end{table}

\begin{table}[ht]
\centering
{\fontsize{7}{10}\selectfont
\caption{\texttt{garber-2001} $(n,p,c)=(66, 4\,554, 4)$}
\label{tab:garber_complet}
\begin{tabular}{llllll}
\toprule
 Model & \ARI & \CH & \Sil&  Time (s) \\
\midrule
\textcolor{blue}{KMeans}& 0.248 $\pm$ 0.044 & 5.606 $\pm$ 1.264 & \bfseries 0.333 $\pm$ 0.000 &  \bfseries 0.056 $\pm$ 0.011 \\
CoHiRF & \bfseries 0.292 $\pm$ 0.034 & \bfseries 5.993 $\pm$ 0.122 & \underline{0.329 $\pm$ 0.006} &   \underline{0.245 $\pm$ 0.118} \\
\RCoHiRF & \underline{0.260 $\pm$ 0.053} & \underline{5.974 $\pm$ 0.132} & 0.327 $\pm$ 0.006 &   0.336 $\pm$ 0.130 \\
\midrule
\textcolor{blue}{\KK}& \underline{0.054 $\pm$ 0.021} & 1.315 $\pm$ 0.099 & 0.035 $\pm$ 0.024 &   \bfseries 0.180 $\pm$ 0.058 \\
CoHiRF & \underline{0.054 $\pm$ 0.015} & \bfseries 1.992 $\pm$ 0.527 & \underline{0.046 $\pm$ 0.067} &   \underline{0.555 $\pm$ 0.225} \\
\RCoHiRF & \bfseries 0.147 $\pm$ 0.056 & \underline{1.862 $\pm$ 0.546} & \bfseries 0.055 $\pm$ 0.018 &   0.639 $\pm$ 0.806 \\
\midrule
\textcolor{blue}{DBSCAN} & \bfseries \underline{0.000 $\pm$ 0.000} & \bfseries \underline{0.000 $\pm$ 0.000} & \bfseries \underline{-1.000 $\pm$ 0.000} &  \bfseries 0.100 $\pm$ 0.007 \\
\CoHiRF & \bfseries \underline{0.000 $\pm$ 0.000} & \bfseries \underline{0.000 $\pm$ 0.000} & \bfseries \underline{-1.000 $\pm$ 0.000} &  \underline{1.974 $\pm$ 0.138} \\
\RCoHiRF  & \bfseries \underline{0.000 $\pm$ 0.000} & \bfseries \underline{0.000 $\pm$ 0.000} & \bfseries \underline{-1.000 $\pm$ 0.000} &  3.419 $\pm$ 0.241 \\
\midrule
\textcolor{blue}{SC-SRGF}& 0.204 $\pm$ 0.035 & 4.966 $\pm$ 0.066 & 0.059 $\pm$ 0.003 &   \bfseries 0.553 $\pm$ 0.291 \\
CoHiRF(1R) & \underline{0.235 $\pm$ 0.013} & \underline{4.994 $\pm$ 0.030} & \underline{0.060 $\pm$ 0.002} &  \underline{1.066 $\pm$ 0.814} \\
CoHiRF(2R) & \bfseries 0.454 $\pm$ 0.034 & \bfseries 4.995 $\pm$ 0.055 & \bfseries 0.229 $\pm$ 0.097 &  1.700 $\pm$ 1.500 \\
\midrule
\bottomrule
\end{tabular}
}
\end{table}


\begin{table}[ht]
\centering
\begin{footnotesize}
\caption{\texttt{shuttle} $(n,p,c)=(58\,000, 10, 7)$.}
\label{tab:shuttle_complet}
\begin{tabular}{llllll}
\toprule
 Model & \ARI & \CH & \Sil&  Time (s) \\
\midrule
  \textcolor{blue}{KMeans} & \underline{0.608 $\pm$ 0.000} & \bfseries 30845.871 $\pm$ 297.412 & \underline{0.960 $\pm$ 0.009} &   \bfseries 0.093 $\pm$ 0.053 \\
  CoHiRF & 0.579 $\pm$ 0.018 & 19657.664 $\pm$ 1672.913 & 0.947 $\pm$ 0.025 &  15.860 $\pm$ 5.240 \\
 \RCoHiRF & \bfseries 0.646 $\pm$ 0.025 & \underline{20315.311 $\pm$ 180.882} & \bfseries 0.969 $\pm$ 0.005 &  20.328 $\pm$ 3.065 \\
\midrule

  \textcolor{blue}{\KK}& 0.250 $\pm$ 0.084 & \underline{11742.148 $\pm$ 2434.781} & 0.390 $\pm$ 0.010 &  \bfseries 1.701 $\pm$ 0.241 \\
 \CoHiRF & \bfseries 0.581 $\pm$ 0.114 & 11265.563 $\pm$ 3001.593 & \underline{0.690 $\pm$ 0.241} &   19.815 $\pm$ 12.709 \\
\RCoHiRF & \underline{0.553 $\pm$ 0.074} & \bfseries 13994.981 $\pm$ 2466.253 & \bfseries 0.856 $\pm$ 0.079 &  32.912 $\pm$ 17.803 \\
\midrule

 \textcolor{blue}{DBSCAN} & 0.709 $\pm$ 0.013 & \bfseries 7504.082 $\pm$ 12.888 & \bfseries 0.948 $\pm$ 0.001 & 36.204 $\pm$ 18.613 \\
\CoHiRF & 0.704 $\pm$ 0.016 & \underline{5529.149 $\pm$ 9498.927} & 0.923 $\pm$ 0.032 &  62.587 $\pm$ 31.710 \\
  \RCoHiRF & 0.696 $\pm$ 0.015 & 1291.726 $\pm$ 583.649 & \underline{0.938 $\pm$ 0.032} &  63.722 $\pm$ 28.394 \\
  \CoHiRFb  & \bfseries 0.723 $\pm$ 0.053 & 1352.078 $\pm$ 307.578 & 0.851 $\pm$ 0.047 &   \bfseries 13.094 $\pm$ 6.924 \\
  \RCoHiRFb & \underline{0.710 $\pm$ 0.055} & 1191.660 $\pm$ 136.736 & 0.916 $\pm$ 0.030 & \underline{13.463 $\pm$ 7.998} \\
\midrule

 \textcolor{blue}{SC-SRGF} & \multicolumn{4}{c}{\textit{infeasible (under our memory limit)}} \\
 \CoHiRFb(1R) & \bfseries \underline{0.456 $\pm$ 0.017} & \bfseries \underline{7317.554 $\pm$ 686.449} & \bfseries \underline{0.573 $\pm$ 0.003} & \bfseries \underline{286.274 $\pm$ 63.310} \\

\bottomrule
\end{tabular}
\end{footnotesize}
\end{table}


\begin{table}[ht]
\centering
\begin{footnotesize}
\caption{\texttt{coil-20} $(n,p,c)=(1\,440, 1\,025, 20)$.}
\label{tab:coil_complet}
\begin{tabular}{llllll}
\toprule
 Model & \ARI & \CH & \Sil&  Time (s) \\
\midrule
   \textcolor{blue}{KMeans} & \bfseries 0.664 $\pm$ 0.005 & \bfseries 288.095 $\pm$ 0.000 & \bfseries 0.225 $\pm$ 0.002 &   \bfseries 0.099 $\pm$ 0.050 \\
  \CoHiRF & \underline{0.335 $\pm$ 0.015} & 288.091 $\pm$ 0.001 & 0.176 $\pm$ 0.004 &   \underline{0.327 $\pm$ 0.119} \\
  \RCoHiRF & 0.244 $\pm$ 0.021 & \underline{288.092 $\pm$ 0.002} & \underline{0.181 $\pm$ 0.012} & 0.726 $\pm$ 0.878 \\
\midrule

   \textcolor{blue}{\KK}& 0.004 $\pm$ 0.003 & 1.795 $\pm$ 0.427 & \underline{0.001 $\pm$ 0.000} &  \bfseries 0.131 $\pm$ 0.046 \\
  \CoHiRF & \bfseries 0.018 $\pm$ 0.028 & \underline{2.606 $\pm$ 2.107} & \bfseries 0.004 $\pm$ 0.001 & 0.871 $\pm$ 1.071 \\
  \RCoHiRF & \underline{0.012 $\pm$ 0.015} & \bfseries 11.301 $\pm$ 13.254 & \bfseries 0.004 $\pm$ 0.001 &  \underline{0.472 $\pm$ 0.113} \\
\midrule

   \textcolor{blue}{DBSCAN} & 0.143 $\pm$ 0.000 & \underline{56.906 $\pm$ 0.000} & -0.005 $\pm$ 0.000 & \bfseries 0.088 $\pm$ 0.009 \\
  \CoHiRF & \bfseries 0.725 $\pm$ 0.021 & 52.067 $\pm$ 78.175 & \underline{0.180 $\pm$ 0.004} &   \underline{1.215 $\pm$ 0.563} \\
   \RCoHiRF  & \underline{0.721 $\pm$ 0.024} & \bfseries 109.584 $\pm$ 199.160 & \bfseries 0.190 $\pm$ 0.004 &  1.379 $\pm$ 0.485 \\
\midrule

   \textcolor{blue}{SC-SRGF}& \bfseries 0.775 $\pm$ 0.005 & \underline{83.629 $\pm$ 1.490} & \bfseries 0.161 $\pm$ 0.000 &  89.340 $\pm$ 46.321 \\
   \CoHiRF(1R) & 0.164 $\pm$ 0.013 & 73.762 $\pm$ 6.568 & \underline{0.121 $\pm$ 0.005} &  \bfseries 57.324 $\pm$ 75.007 \\
    \CoHiRF(2R) & \underline{0.200 $\pm$ 0.014} & \bfseries 88.969 $\pm$ 4.455 & 0.111 $\pm$ 0.028 &  \underline{65.973 $\pm$ 47.384} \\
\bottomrule
\end{tabular}
\end{footnotesize}
\end{table}

\end{document}